\documentclass[runningheads]{llncs}
\usepackage{graphicx}
\usepackage{amsmath,amssymb} % define this before the line numbering.
\usepackage{booktabs}        % professional-quality tables
\usepackage[                % hyperlinks
   pagebackref=false,
   breaklinks=true,
   letterpaper=true,
   colorlinks,
   bookmarks=false]{hyperref}
\usepackage{pifont}
\usepackage{xcolor}
\usepackage{cleveref}
\usepackage{lipsum}
\usepackage{caption}
\usepackage{subcaption}
\usepackage{capt-of}
\usepackage{cite}
\usepackage{standalone}
\usepackage{tikz}
\usepackage{enumitem}
\usepackage{picins}
\usepackage{multirow}
\usepackage{floatrow}
\usepackage{etex}
\usetikzlibrary{backgrounds}

\newcommand{\cmark}{\ding{51}}

\usepackage[
   width=122mm,
   left=12mm,
   paperwidth=146mm,
   height=193mm,
   top=12mm,
   paperheight=217mm]{geometry}

\usepackage{upgreek}
\usepackage{bm}
\makeatletter
\newcommand{\bfgreek}[1]{\bm{\@nameuse{#1}}}
\makeatother

\begin{document}

\def\eg{\emph{e.g.}}
\def\Eg{\emph{E.g.}}
\def\etal{\emph{et al.}}
\def\ypad{\phantom{k[}}

\newcommand{\gt}[1]{examples/ctest10k-supp/gt/#1}
\newcommand{\gs}[1]{examples/ctest10k-supp/grayscale/#1}
\newcommand{\mm}[1]{examples/ctest10k-supp/output/#1}

\newcommand\filler{\textcolor{gray}{\lipsum[66]}}
\newcommand\placeholder[1]{\textcolor{black!60!red}{[#1]}}
\newcommand\todo[1]{\textcolor{red}{\{#1\}}}

\def\bx{\mathbf{x}}
\def\by{\mathbf{y}}
\def\bz{\mathbf{z}}
\def\bp{\mathbf{p}}
\def\bt{\mathbf{t}}
\def\bb{\mathbf{b}}
\def\bW{\mathbf{W}}

\newcommand{\gscomment}[1]{\textcolor{blue}{#1}}
\newcommand{\glcomment}[1]{\textcolor{green}{#1}}
\newcommand{\mmcomment}[1]{\textcolor{tan}{#1}}

\pagestyle{headings}
\mainmatter
\def\ECCV16SubNumber{860}  % Insert your submission number here

\title{\large{Learning Representations for Automatic Colorization}}

\titlerunning{Learning Representations for Automatic Colorization}

\authorrunning{Larsson, Maire, Shakhnarovich}

\author{
    Gustav Larsson$^1$ \and
    Michael Maire$^2$ \and
    Gregory Shakhnarovich$^2$
}
\institute{
    $^1$University of Chicago \quad
    $^2$Toyota Technological Institute at Chicago\\
    {\tt \small larsson@cs.uchicago.edu, \{mmaire,greg\}@ttic.edu}
}

\maketitle

%auto-ignore
\begin{abstract}
We develop a fully automatic image colorization system.  Our approach
leverages recent advances in deep networks, exploiting both low-level and
semantic representations.  As many scene elements naturally appear according to
multimodal color distributions, we train our model to predict per-pixel color
histograms.  This intermediate output can be used to automatically generate a
color image, or further manipulated prior to image formation.  On both fully
and partially automatic colorization tasks, we outperform existing methods.
We also explore colorization as a vehicle for self-supervised visual
representation learning.
% keywords can be omitted from review copy
%\keywords{
%  Automatic colorization, convolutional neural networks, hypercolumns, deep learning.
%}
\end{abstract}

%\begin{figure}[h]
   %\RawFloats
   %\centering
\begin{center}
   \begin{tabular}{cccccc}
      \includegraphics[height=.115\textwidth]{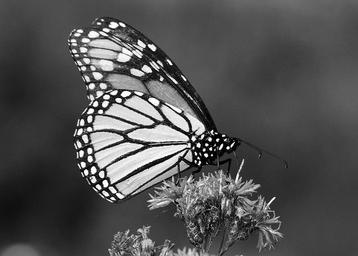}&
      \includegraphics[height=.115\textwidth]{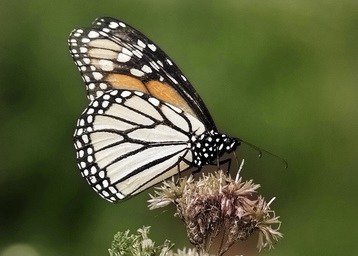}&
      \includegraphics[height=.115\textwidth]{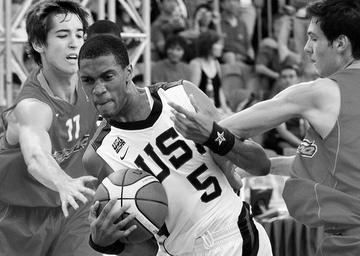}&
      \includegraphics[height=.115\textwidth]{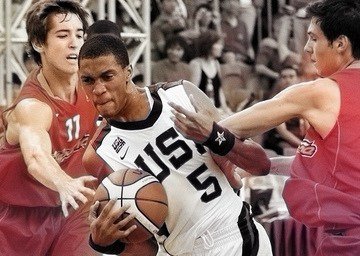}&
      \includegraphics[height=.115\textwidth]{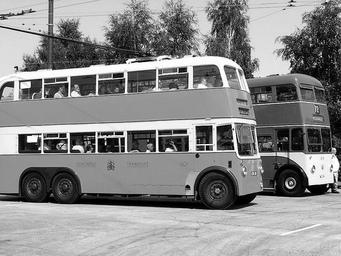}&
      \includegraphics[height=.115\textwidth]{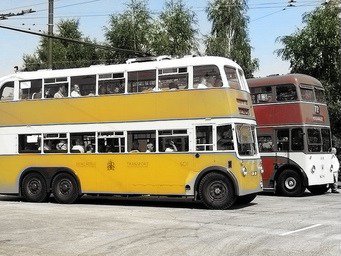}
   \end{tabular}
   \captionof{figure}{
      \small
      Our automatic colorization of grayscale input; more examples in
      Figs.~\ref{fig:ctest10k-examples}~and~\ref{fig:ctest10k-more-examples}.
   }
   \label{fig:teaser}
\end{center}
%\end{figure}

%auto-ignore
\section{Introduction}
\label{sec:intro}

Colorization of grayscale images is a simple task for the human imagination.
A human need only recall that sky is blue and grass is green; for many objects,
the mind is free to hallucinate several plausible colors.  The high-level
comprehension required for this process is precisely why the development of
fully automatic colorization algorithms remains a challenge.  Colorization is
thus intriguing beyond its immediate practical utility in graphics
applications.  Automatic colorization serves as a proxy measure for visual
understanding.  Our work makes this connection explicit; we unify a
colorization pipeline with the type of deep neural architectures driving
advances in image classification and object detection.

Both our technical approach and focus on fully automatic results depart from
past work.  Given colorization's importance across multiple applications
(\eg~historical photographs and videos~\cite{tsaftaris2014novel}, artist
assistance~\cite{sykora2004unsupervised,qu2006manga}), much research strives
to make it cheaper and less time-consuming~\cite{
   welsh2002transferring,
   levin2004colorization,
   irony2005colorization,
   charpiat2008automatic,
   morimoto2009automatic,
   chia2011semantic,
   gupta2012image,
   deshpande2015learning,
   cheng2015deep}.
However, most methods still require some level of user input~\cite{
   levin2004colorization,
   sapiro2005inpainting,
   irony2005colorization,
   charpiat2008automatic,
   chia2011semantic,
   gupta2012image}.
Our work joins the relatively few recent efforts on fully automatic
colorization~\cite{morimoto2009automatic,deshpande2015learning,cheng2015deep}.
Some~\cite{deshpande2015learning, cheng2015deep} show promising results on
typical scenes (\eg~landscapes), but their success is limited on complex
images with foreground objects.

\begin{figure}[!t]
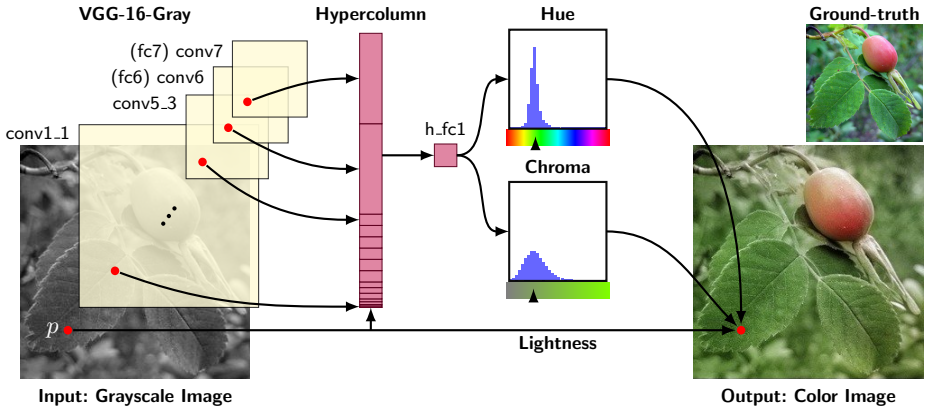

   \RawFloats
   \hspace{-0.35cm}
   \includestandalone[trim=0.1cm 0.1cm 0.1cm 0.1cm]{figures/overview}
   \vspace{-0.2cm}
   \caption{
   \small
   \textbf{System overview.}
      We process a grayscale image through a deep convolutional
      architecture~(VGG)~\cite{vgg16} and take spatially localized multilayer
      slices (hypercolumns)~\cite{
         MYP:ACCV:2014, mostajabi2015feedforward ,hariharan2015hypercolumns},
      as per-pixel descriptors.  We train our system end-to-end for the task
      of predicting hue and chroma distributions for each pixel $p$ given its
      hypercolumn descriptor.  These predicted distributions determine color
      assignment at test time.
   }
   \label{fig:schematic}
\end{figure}

At a technical level, existing automatic colorization methods often employ
a strategy of finding suitable reference images and transferring their
color onto a target grayscale image~\cite{morimoto2009automatic,
deshpande2015learning}.  This works well if sufficiently similar reference
images can be found, but is difficult for unique grayscale input images.  Such
a strategy also requires processing a large repository of reference images at
test time.  In contrast, our approach is free of database search and fast at
test time.  Section~\ref{sec:related} provides a complete view of prior
methods, highlighting differences.

Our approach to automatic colorization converts two intuitive observations
into design principles.  First, semantic information matters.  In order to
colorize arbitrary images, a system must interpret the semantic composition
of the scene (what is in the image: faces, cars, plants, \ldots) as well as
localize objects (where things are).  Deep convolutional neural networks (CNNs)
can serve as tools to incorporate semantic parsing and localization into a
colorization system.

Our second observation is that while some scene elements can be assigned a
single color with high confidence, others (\eg~clothes or cars) may draw from
many suitable colors.  Thus, we design our system to predict a color histogram,
instead of a single color, at every image location.  Figure~\ref{fig:schematic}
sketches the CNN architecture we use to connect semantics with color
distributions by exploiting features across multiple abstraction levels.
Section~\ref{sec:method} provides details.

Section~\ref{sec:experiments} experimentally validates our algorithm against
competing methods~\cite{welsh2002transferring,deshpande2015learning} in two
settings:  fully (grayscale input only) and partially (grayscale input with
reference global color histogram) automatic colorization.  Across every
metric and dataset~\cite{xiao2010sun,patterson2014sun,imagenet}, our method
achieves the best performance.  Our system's fully automatic output is superior
to that of prior methods relying on additional information such as reference
images or ground-truth color histograms.  To ease the comparison burden for
future research, we propose a new colorization benchmark on
ImageNet~\cite{imagenet}.  We also experiment with colorization itself
as an objective for learning visual representations from scratch, thereby
replacing use of ImageNet pretraining in a traditional semantic labeling task.

Section~\ref{sec:final} summarizes our contributions: (1) a novel
technical approach to colorization, bringing semantic knowledge to bear using
CNNs, and modeling color distributions;  (2) state-of-the-art performance
across fully and partially automatic colorization tasks; (3) a new ImageNet
colorization benchmark; (4) proof of concept on colorization for
self-supervised representation learning.

%auto-ignore
\section{Related work}
\label{sec:related}

Previous colorization methods broadly fall into three categories:
scribble-based~\cite{
   levin2004colorization,
   huang2005adaptive,
   qu2006manga,
   yatziv2006fast,
   luan2007natural},
transfer~\cite{
   welsh2002transferring,
   irony2005colorization,
   tai2005local,
   charpiat2008automatic,
   morimoto2009automatic,
   chia2011semantic,
   gupta2012image},
and automatic direct prediction~\cite{
   deshpande2015learning,cheng2015deep}.

\emph{Scribble-based} methods, introduced by
Levin~\etal~\cite{levin2004colorization}, require manually specifying
desired colors of certain regions.  These scribble colors are propagated under
the assumption that adjacent pixels with similar luminance should have similar
color, with the optimization relying on Normalized Cuts~\cite{
shi2000normalized}.  Users can interactively refine results via additional
scribbles.  Further advances extend similarity to
texture~\cite{qu2006manga,luan2007natural}, and exploit edges to reduce color
bleeding~\cite{huang2005adaptive}.

\emph{Transfer-based} methods rely on availability of related \emph{reference}
image(s), from which color is transferred to the target grayscale image.
Mapping between source and target is established automatically, using
correspondences between local descriptors~\cite{welsh2002transferring,
charpiat2008automatic,morimoto2009automatic}, or in combination with manual
intervention~\cite{irony2005colorization,chia2011semantic}.
Excepting~\cite{morimoto2009automatic}, reference image selection is at
least partially manual.

In contrast to these method families, our goal is \emph{fully automatic}
colorization.  We are aware of two recent efforts in this direction.
Deshpande~\etal~\cite{deshpande2015learning} colorize an entire image by
solving a linear system.  This can be seen as an extension of patch-matching
techniques~\cite{welsh2002transferring}, adding interaction terms for spatial
consistency.  Regression trees address the high-dimensionality of the system.
Inference requires an iterative algorithm.  Most of the experiments are
focused on a dataset (SUN-6) limited to images of a few scene classes, and
best results are obtained when the scene class is known at test time.  They
also examine another partially automatic task, in which a desired global color
histogram is provided.

The work of Cheng~\etal~\cite{cheng2015deep} is perhaps most related to ours.
It combines three levels of features with increasing receptive field:
   the raw image patch,
   DAISY features~\cite{tola2008fast}, and
   semantic features~\cite{long2015fully}.
These features are concatenated and fed into a three-layer fully connected
neural network trained with an $L_2$ loss.  Only this last component is
optimized; the feature representations are fixed.

Unlike~\cite{deshpande2015learning,cheng2015deep}, our system
does not rely on hand-crafted features, is trained end-to-end, and treats
color prediction as a histogram estimation task rather than as regression.
Experiments in Section~\ref{sec:experiments} justify these principles by
demonstrating performance superior to the best reported by~\cite{
deshpande2015learning,cheng2015deep} across all regimes.
%But first, we elaborate the specifics of our architecture and algorithm.

% Iizuka & Simo-Serra et al.
% * uses separate global and local networks - we think a hypercolumn is a more natural architecture for coming low- to high-level features
% * trains classification and colorization jointly - may bias results too much to classes
% * Uses L2 regression

% Zhang et al. (them / us)
% * up-convolutional / fully convolutional
% * deep supervision / hypercolumn

Two concurrent efforts also present feed-forward networks trained end-to-end
for colorization.  Iizuka \& Simo-Serra~\etal~\cite{IizukaSIGGRAPH2016} propose
a network that concatenates two separate paths, specializing in global and
local features, respectively.  This concatenation can be seen as a two-tiered
hypercolumn; in comparison, our 16-layer hypercolumn creates a continuum
between low- and high-level features.  Their network is trained jointly for
classification (cross-entropy) and colorization ($L_2$ loss in Lab).  We
initialize, but do not anchor, our system to a classification-based network,
allowing for fine-tuning of colorization on unlabeled datasets.

Zhang~\etal~\cite{zhang2016colorful} similarly propose predicting color
histograms to handle multi-modality.  Some key differences include their usage
of up-convolutional layers, deep supervision, and dense training.  In
comparison, we use a fully convolutional approach, with deep supervision
implicit in the hypercolumn design, and, as Section~\ref{sec:method} describes,
memory-efficient training via spatially sparse samples.

%auto-ignore
\section{Method}
\label{sec:method}

We frame the colorization problem as learning a function
$f : \mathcal{X} \to \mathcal{Y}$.  Given a grayscale image patch
$\bx \in \mathcal{X} = [0, 1]^{S \times S}$, $f$ predicts the color $\by \in
\mathcal{Y}$ of its center pixel.  The patch size $S \times S$ is the
receptive field of the colorizer.  The output space $\mathcal{Y}$ depends
on the choice of color parameterization.  We implement $f$ according to the
neural network architecture diagrammed in Figure~\ref{fig:schematic}.

Motivating this strategy is the success of similar architectures for
semantic segmentation~\cite{
   farabet2013learning,
   long2015fully,
   chen2014semantic,
   hariharan2015hypercolumns,
   mostajabi2015feedforward}
and edge detection~\cite{
   MYP:ACCV:2014,
   GL:ACCV:2014,
   BST:CVPR:2015,
   SWWBZ:CVPR:2015,
   xie2015holistically}.
Together with colorization, these tasks can all be viewed as image-to-image
prediction problems, in which a value is predicted for each input pixel.
Leading methods commonly adapt deep convolutional neural networks
pretrained for image classification~\cite{imagenet,vgg16}.  Such classification
networks can be converted to \emph{fully convolutional} networks that
produce output of the same spatial size as the input, \eg~using the
shift-and-stitch method~\cite{long2015fully} or the more efficient {\em\`a
trous} algorithm~\cite{chen2014semantic}.  Subsequent training with a
task-specific loss fine-tunes the converted network.

Skip-layer connections, which directly link low- and mid-level features to
prediction layers, are an architectural addition beneficial for many
image-to-image problems.  Some methods implement skip connections
directly through concatenation layers~\cite{long2015fully,chen2014semantic},
while others equivalently extract per-pixel descriptors by reading localized
slices of multiple layers~\cite{MYP:ACCV:2014,mostajabi2015feedforward,
hariharan2015hypercolumns}.  We use this latter strategy and adopt the
recently coined \emph{hypercolumn} terminology~\cite{hariharan2015hypercolumns}
for such slices.

Though we build upon these ideas, our technical approach innovates on two
fronts.  First, we integrate domain knowledge for colorization, experimenting
with output spaces and loss functions.  We design the network output to serve
as an intermediate representation, appropriate for direct or biased sampling.
We introduce an energy minimization procedure for optionally biasing sampling
towards a reference image.  Second, we develop a novel and efficient
computational strategy for network training that is widely applicable to
hypercolumn architectures.

\subsection{Color spaces}
\label{sec:method_color}

We generate training data by converting color images to grayscale according to
$L = \frac{R+G+B}{3}$.  This is only one of many desaturation options and
chosen primarily to facilitate comparison with Deshpande~\etal~\cite{
deshpande2015learning}.  For the representation of color predictions, using
RGB is overdetermined, as lightness $L$ is already known.  We instead consider
output color spaces with $L$ (or a closely related quantity) conveniently
appearing as a separate pass-through channel:
\begin{itemize}
   \item{
      \textbf{Hue/chroma}.
      Hue-based spaces, such as HSL, can be thought of as a color cylinder,
      with angular coordinate $H$ (hue), radial distance $S$ (saturation),
      and height $L$ (lightness).  The values of $S$ and $H$ are unstable
      at the bottom (black) and top (white) of the cylinder.  HSV
      describes a similar color cylinder which is only unstable at the bottom.
      However, $L$ is no longer one of the channels.  We wish to avoid both
      instabilities and still retain $L$ as a channel.  The solution is a
      color bicone, where chroma ($C$) takes the place of saturation. 
      % This
      %color space only has instability in $H$ around $C = 0$; this is inherent
      %to all hue-based color spaces and we show how to overcome this when
      %designing the loss function.
      Conversion to HSV is given by
         $V\,=\,L\,+\,\frac{C}{2},\; S\,=\,\frac{C}{V}$.
   }
   \item{
      \textbf{Lab} and $\bfgreek{alpha}\bfgreek{beta}$.
      Lab (or L*a*b) is designed to be perceptually linear.  The color vector
      $(a,b)$ defines a Euclidean space where the distance to the origin
      determines chroma.  Deshpande~\etal~\cite{deshpande2015learning} use a
      color space somewhat similar to Lab, denoted ``ab''. To differentiate,
      we call their color space $\alpha\beta$.
   }
\end{itemize}

\subsection{Loss}
\label{sec:method_loss}

For any output color representation, we require a loss function for measuring
prediction errors.  A first consideration, also used in~\cite{cheng2015deep},
is $L_2$ regression in Lab:
\begin{equation}
   \label{eq:loss-l2}
    L_\mathrm{reg}(\bx, \by) = \| f(\bx) - \by \|^2
\end{equation}
where $\mathcal{Y} = \mathbb{R}^2$ describes the $(a, b)$ vector space.
However, regression targets do not handle multimodal color distributions well.
To address this, we instead predict distributions over a set of color bins,
a technique also used in~\cite{charpiat2008automatic}:
\begin{equation}
   \label{eq:loss-hist}
   L_\mathrm{hist}(\bx, \by) = D_\mathrm{KL}( \by \|f(\bx))
\end{equation}
where $\mathcal{Y} = [0, 1]^K$ describes a histogram over $K$ bins, and
$D_\mathrm{KL}$ is the KL-divergence.  The ground-truth histogram $\by$ is set
as the empirical distribution in a rectangular region of size $R$ around the
center pixel.  Somewhat surprisingly, our experiments see no benefit to
predicting smoothed histograms, so we simply set $R = 1$.  This makes $\by$ a
one-hot vector and Equation~\eqref{eq:loss-hist} the log loss.  For histogram
predictions, the last layer of neural network $f$ is always a softmax.

There are several choices of how to bin color space.  We bin the Lab axes by
evenly spaced Gaussian quantiles ($\mu = 0, \sigma = 25$).  They can be
encoded separately for $a$ and $b$ (as marginal distributions), in which case
our loss becomes the sum of two separate terms defined by Equation~\eqref{eq:loss-hist}.
They can also be encoded as a joint distribution over $a$ and $b$, in which
case we let the quantiles form a 2D grid of bins.  In our experiments, we set
$K = 32$ for marginal distributions and $K = 16 \times 16$ for joint.  We
determined these numbers, along with $\sigma$, to offer a good compromise
of output fidelity and output complexity.

For hue/chroma, we only consider marginal distributions and bin axes uniformly
in $[0,1]$.  Since hue becomes unstable as chroma approaches zero, we add a
sample weight to the hue based on the chroma:
\begin{equation} \label{eq:loss-hc}
   L_\mathrm{hue/chroma}(\bx, \by) =
      D_\mathrm{KL}( \by_\mathrm{C}\|f_\mathrm{C}(\bx) ) +
      \lambda_H y_\mathrm{C} D_\mathrm{KL}(\by_\mathrm{H}\|f_\mathrm{H}(\bx))
\end{equation}
where $\mathcal{Y} = [0,1]^{2 \times K}$ and $y_C \in [0,1]$ is the sample
pixel's chroma.  We set $\lambda_H = 5$, roughly the inverse expectation of
$y_\mathrm{C}$, thus equally weighting hue and chroma.

\subsection{Inference}
\label{sec:method_inference}

Given network $f$ trained according to a loss function in the previous section,
we evaluate it at every pixel $n$ in a test image: $\hat{\by}_n = f(\bx_n)$.
For the $L_2$ loss, all that remains is to combine each $\hat{\by}_n$ with the
respective lightness and convert to RGB.  With histogram predictions, we
consider options for inferring a final color:

\begin{itemize}
   \item{
      \textbf{Sample}
      Draw a sample from the histogram.  If done per pixel, this may create
      high-frequency color changes in areas of high-entropy histograms.
   }
   \item{
      \textbf{Mode}
      Take the $\arg\max_k \hat{y}_{n,k}$ as the color.  This can create
      jarring transitions between colors, and is prone to vote
      splitting for proximal centroids.
   }
   \item{
      \textbf{Median}
      Compute cumulative sum of $\hat{\by}_n$ and use linear interpolation to
      find the value at the middle bin.  Undefined for circular
      histograms, such as hue.
   }
   \item{
      \textbf{Expectation}
      Sum over the color bin centroids weighted by the histogram.
   }
\end{itemize}
For Lab output, we achieve the best qualitative and quantitative results
using expectations.  For hue/chroma, the best results are achieved by taking
the median of the chroma.  Many objects can appear both with and without
chroma, which means $C = 0$ is a particularly common bin.  This mode draws the
expectation closer to zero, producing less saturated images.  As for hue, since
it is circular, we first compute the complex expectation:
\begin{equation}
   z =
      \mathbb{E}_{H \sim f_h(\bx)}[H]
      \triangleq \frac 1K \sum_k [f_h(x)]_k \mathrm{e}^{i\theta_k}, \quad
      \theta_k = 2\pi \frac{k + 0.5}{K}
\end{equation}
We then set hue to the argument of $z$ remapped to lie in $[0,1)$.

\parpic[r][t]{ \includegraphics{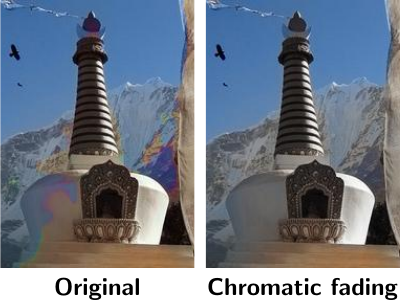} }

In cases where the estimate of the chroma is high and $z$ is close to zero,
the instability of the hue can create artifacts.
%(Figure~\ref{fig:chromatic-fading}).
A simple, yet effective, fix is chromatic fading: downweight the chroma if the
absolute value of $z$ is too small.  We thus re-define the predicted chroma by
multiplying it by a factor of $\max(\eta^{-1}|z|, 1)$.
In our experiments, we set $\eta = 0.03$ (obtained via cross-validation).

\subsection{Histogram transfer from ground-truth}
\label{sec:method_transfer}

So far, we have only considered the fully automatic color inference task.
Deshpande~\etal~\cite{deshpande2015learning}, test a separate task where
the ground-truth histogram in the two non-lightness color channels of the
original color image is made available.\footnote{
   Note that if the histogram of the $L$ channel were available, it would be
   possible to match lightness to lightness exactly and thus greatly narrow
   down color placement.
}
In order to compare, we propose two histogram transfer methods.  We refer to
the predicted image as the \emph{source} and the ground-truth image as the
\emph{target}.

~\\\noindent
\textbf{Lightness-normalized quantile matching}.
\label{sec:energy}
Divide the RGB representation of both source and target by their respective
lightness.  Compute marginal histograms over the resulting three color
channels.  Alter each source histogram to fit the corresponding target
histogram by quantile matching, and multiply by lightness.  Though it does not
exploit our richer color distribution predictions, quantile matching beats the
cluster correspondence method of~\cite{deshpande2015learning}
(see Table~\ref{tab:sun6}).

~\\\noindent
\textbf{Energy minimization}.
We phrase histogram matching as minimizing energy:
\begin{equation}\label{eq:energy-min}
   E =
      \frac 1N \sum_n D_\mathrm{KL}(\hat{\by}^*_n \| \hat{\by}_n) +
      \lambda D_{\chi^2}(\langle \hat{\by^*} \rangle, \bt)
\end{equation}
where $N$ is the number of pixels,
$\hat{\by}, \hat{\by}^* \in [0, 1]^{N \times K}$ are the predicted and
posterior distributions, respectively.  The target histogram is denoted by
$\bt \in [0, 1]^K$.  The first term contains unary potentials that anchor
the posteriors to the predictions.  The second term is a symmetric $\chi^2$
distance to promote proximity between source and target histograms.  Weight
$\lambda$ defines relative importance of histogram matching.  We estimate the
source histogram as
   $\langle \hat{\by}^* \rangle = \frac 1N \sum_n \hat{\by}^*_n$.
We parameterize the posterior for all pixels $n$ as:
   $\hat{\by}^*_n = \mathrm{softmax}(\log \hat{\by}_n + \bb)$,
where the vector $\bb \in \mathbb{R}^K$ can be seen as a global bias for each
bin.  It is also possible to solve for the posteriors directly; this does not
perform  better quantitatively and is more prone to introducing artifacts.
We solve for $\bb$ using gradient descent on $E$ and use the resulting
posteriors in place of the predictions.  In the case of marginal
histograms, the optimization is run twice, once for each color channel.

\subsection{Neural network architecture and training}

Our base network is a fully convolutional version of VGG-16~\cite{vgg16} with
two changes: (1) the classification layer ($\texttt{fc8}$) is discarded, and
(2) the first filter layer (\texttt{conv1\_1}) operates on a single intensity
channel instead of mean-subtracted RGB.  We extract a hypercolumn descriptor
for a pixel by concatenating the features at its spatial location in all
layers, from \texttt{data} to \texttt{conv7} (\texttt{fc7}), resulting in a
$12,417$ channel descriptor.  We feed this hypercolumn into a fully connected
layer with $1024$ channels (\texttt{h\_fc1} in Figure~\ref{fig:schematic}), to
which we connect output predictors.

Processing each pixel separately in such manner is quite costly.  We
instead run an entire image through a single forward pass of VGG-16 and
approximate hypercolumns using bilinear interpolation.  Even with such
sharing, densely extracting hypercolumns requires significant memory
($1.7$~GB for $256\times256$ input).

\begin{figure}[!th]
    \begin{center}
    \begin{minipage}[b]{0.156\linewidth}
        \vspace{0pt}
        \begin{center}
            \includegraphics[width=\textwidth]{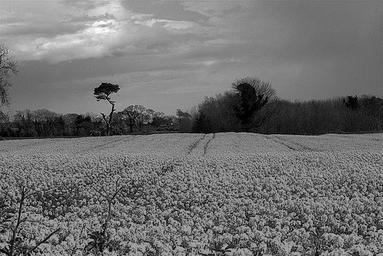}\\
            \includegraphics[width=\textwidth]{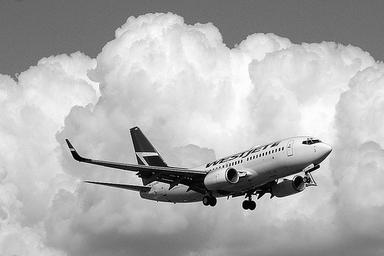}\\
            \includegraphics[width=\textwidth]{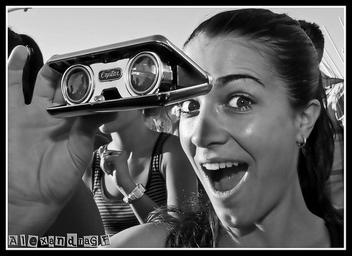}\\
            \includegraphics[width=\textwidth]{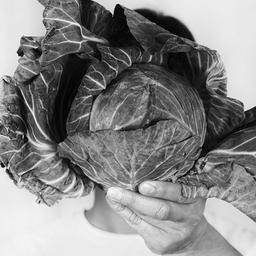}\\
            \includegraphics[width=\textwidth]{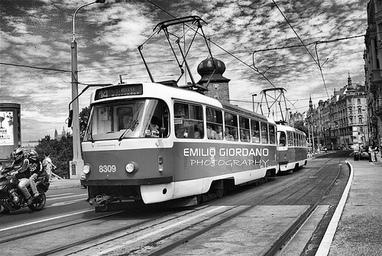}\\
            \includegraphics[width=\textwidth]{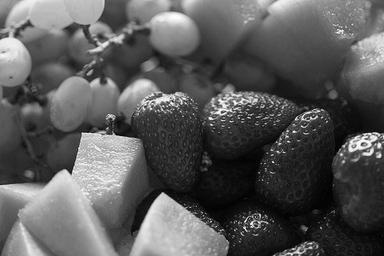}\\
            \scriptsize{\textbf{\textsf{Input}}}
        \end{center}
    \end{minipage}
    \begin{minipage}[b]{0.156\linewidth}
        \vspace{0pt}
        \begin{center}
            \includegraphics[width=\textwidth]{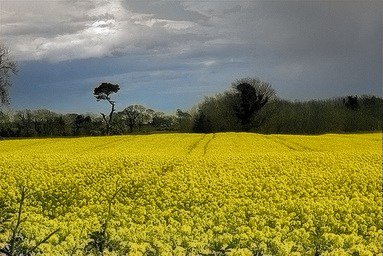}\\
            \includegraphics[width=\textwidth]{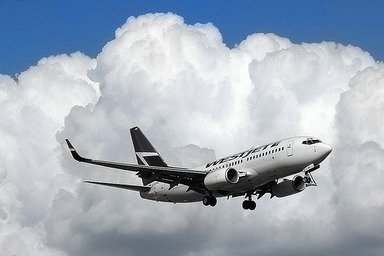}\\
            \includegraphics[width=\textwidth]{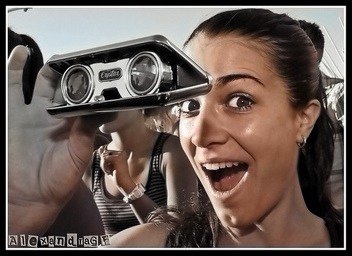}\\
            \includegraphics[width=\textwidth]{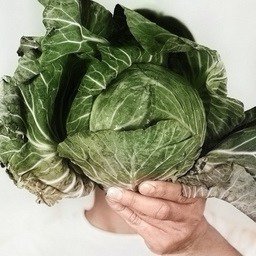}\\
            \includegraphics[width=\textwidth]{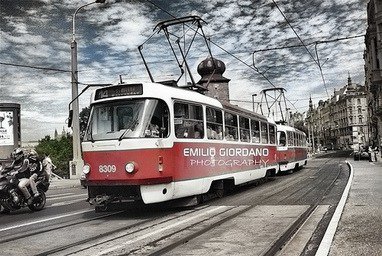}\\
            \includegraphics[width=\textwidth]{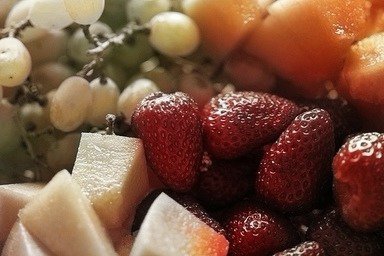}\\
            \scriptsize{\textbf{\textsf{Our Method}}}
        \end{center}
    \end{minipage}
    \begin{minipage}[b]{0.156\linewidth}
        \vspace{0pt}
        \begin{center}
            \includegraphics[width=\textwidth]{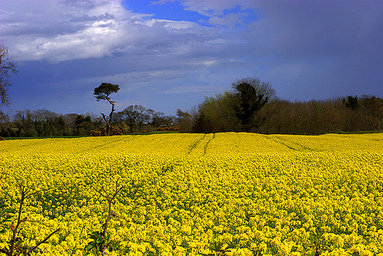}\\
            \includegraphics[width=\textwidth]{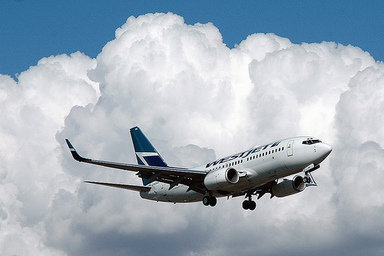}\\
            \includegraphics[width=\textwidth]{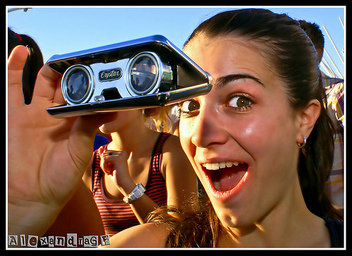}\\
            \includegraphics[width=\textwidth]{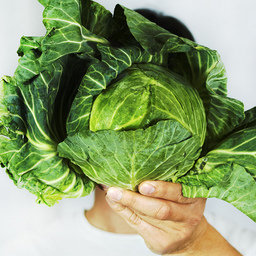}\\
            \includegraphics[width=\textwidth]{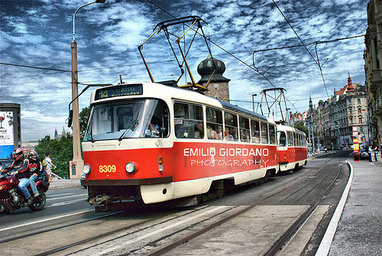}\\
            \includegraphics[width=\textwidth]{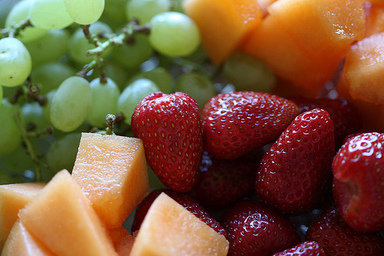}\\
            \scriptsize{\textbf{\textsf{Ground-truth}}}
        \end{center}
    \end{minipage}
    \hfill
    \begin{minipage}[b]{0.156\linewidth}
        \vspace{0pt}
        \begin{center}
            \includegraphics[width=\textwidth]{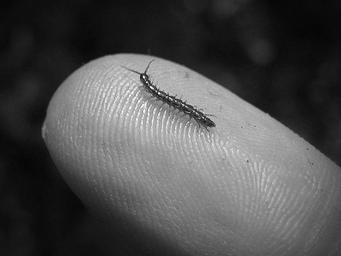}\\
            \vspace{0.0075\linewidth}
            \includegraphics[width=\textwidth]{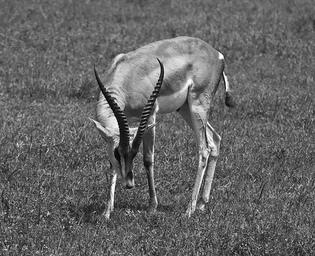}\\
            \vspace{0.0075\linewidth}
            \includegraphics[width=\textwidth]{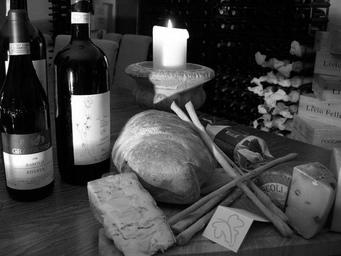}\\
            \vspace{0.0075\linewidth}
            \includegraphics[width=\textwidth]{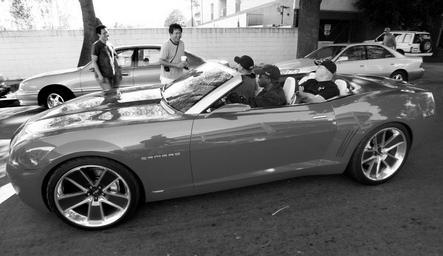}\\
            %\vspace{0.0075\linewidth}
            %\includegraphics[width=\textwidth]{examples/ctest10k/grayscale/00040826}\\
            \vspace{0.0075\linewidth}
            \includegraphics[width=\textwidth]{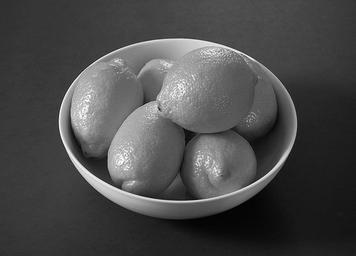}\\
            \vspace{0.0075\linewidth}
            \includegraphics[width=\textwidth]{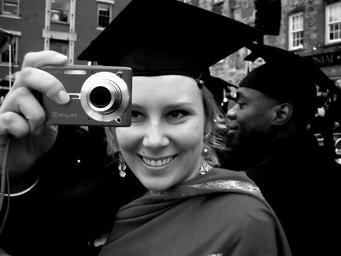}\\
            \scriptsize{\textbf{\textsf{Input}}}
        \end{center}
    \end{minipage}
    \begin{minipage}[b]{0.156\linewidth}
        \vspace{0pt}
        \begin{center}
            \includegraphics[width=\textwidth]{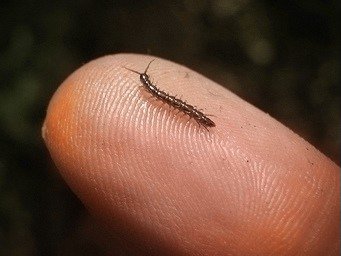}\\
            \vspace{0.0075\linewidth}
            \includegraphics[width=\textwidth]{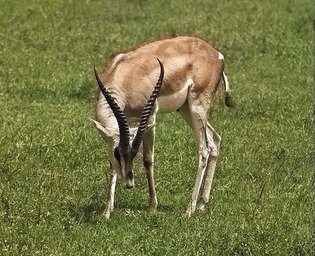}\\
            \vspace{0.0075\linewidth}
            \includegraphics[width=\textwidth]{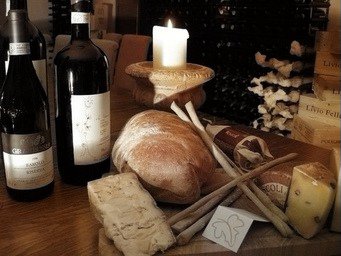}\\
            \vspace{0.0075\linewidth}
            \includegraphics[width=\textwidth]{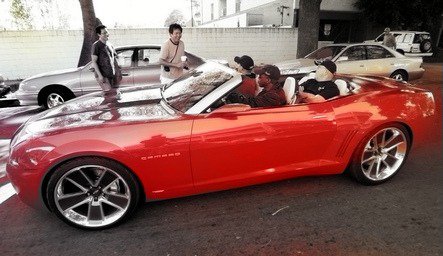}\\
            \vspace{0.0075\linewidth}
            %\includegraphics[width=\textwidth]{examples/ctest10k/good/00040826}\\
            %\vspace{0.0075\linewidth}
            \includegraphics[width=\textwidth]{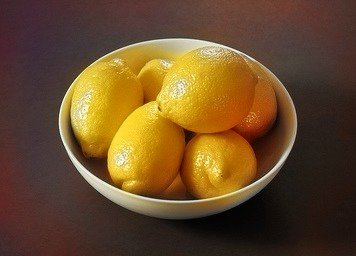}\\
            \vspace{0.0075\linewidth}
            \includegraphics[width=\textwidth]{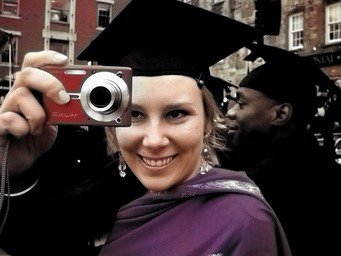}\\
            \scriptsize{\textbf{\textsf{Our Method}}}
        \end{center}
    \end{minipage}
    \begin{minipage}[b]{0.156\linewidth}
        \vspace{0pt}
        \begin{center}
            \includegraphics[width=\textwidth]{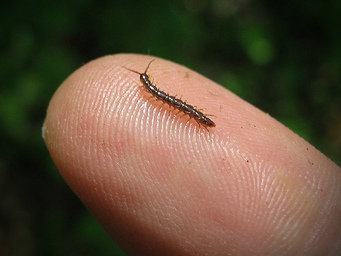}\\
            \vspace{0.0075\linewidth}
            \includegraphics[width=\textwidth]{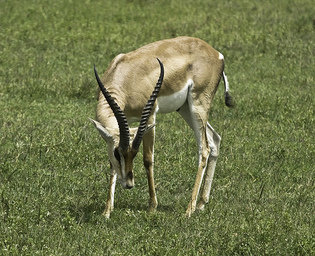}\\
            \vspace{0.0075\linewidth}
            \includegraphics[width=\textwidth]{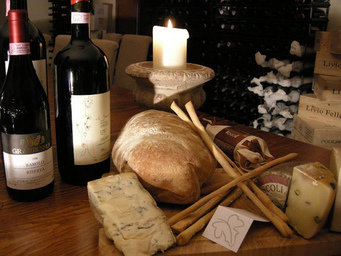}\\
            \vspace{0.0075\linewidth}
            \includegraphics[width=\textwidth]{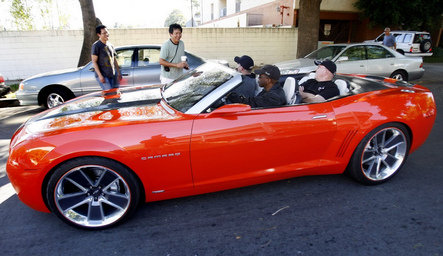}\\
            \vspace{0.0075\linewidth}
            %\includegraphics[width=\textwidth]{examples/ctest10k/good-gt/00040826}\\
            %\vspace{0.0075\linewidth}
            \includegraphics[width=\textwidth]{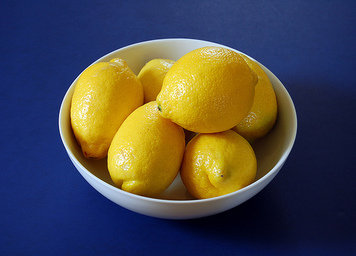}\\
            \vspace{0.0075\linewidth}
            \includegraphics[width=\textwidth]{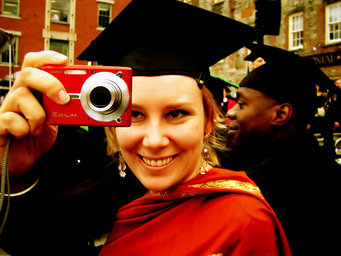}\\
            \scriptsize{\textbf{\textsf{Ground-truth}}}
        \end{center}
    \end{minipage}
    \end{center}
    \caption{\small 
        \textbf{Fully automatic colorization results on ImageNet/ctest10k.}
        Our system reproduces known object color properties (\eg~faces, sky,
        grass, fruit, wood), and coherently picks colors for objects without
        such properties (\eg~clothing).
    }
    \label{fig:ctest10k-examples}
\end{figure}

To fit image batches in memory during training, we instead extract hypercolumns
at only a sparse set of locations, implementing a custom
Caffe~\cite{jia2014caffe} layer to directly compute them.\footnote{
   \url{https://github.com/gustavla/autocolorize}
}
Extracting batches of only $128$ hypercolumn descriptors per input image,
sampled at random locations, provides sufficient training signal.  In the
backward pass of stochastic gradient descent, an interpolated hypercolumn
propagates its gradients to the four closest spatial cells in each layer.
Locks ensure atomicity of gradient updates, without incurring any performance
penalty.  This drops training memory for hypercolumns to only $13$~MB per
image.

We initialize with a version of VGG-16 pretrained on ImageNet, adapting it to
grayscale by averaging over color channels in the first layer and rescaling
appropriately.  Prior to training for colorization, we further fine-tune the
network for one epoch on the ImageNet classification task with grayscale
input.  As the original VGG-16 was trained without batch normalization~\cite{
ioffe2015batch}, scale of responses in internal layers can vary dramatically,
presenting a problem for learning atop their hypercolumn concatenation.
Liu~\etal~\cite{liu2015parsenet} compensate for such variability by applying
layer-wise $L_2$ normalization.  We use the alternative of balancing
hypercolumns so that each layer has roughly unit second moment
($\mathbb{E}[X^2] \approx 1$); Appendix (Section~\ref{sec:rebalance}) provides additional
details.

%auto-ignore
\begin{figure}[!th]
    \begin{center}
    \begin{minipage}[b]{0.157\linewidth}
        \begin{center}
            \includegraphics[width=\textwidth]{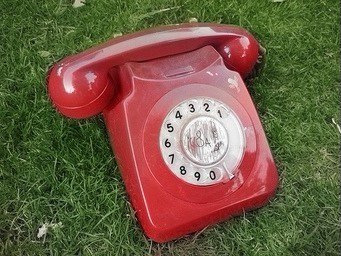}\\
            \includegraphics[width=\textwidth]{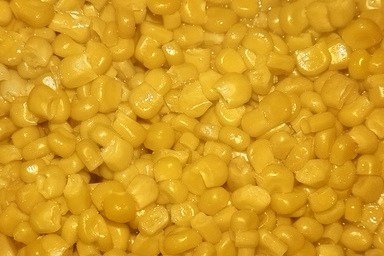}\\
            \includegraphics[width=\textwidth]{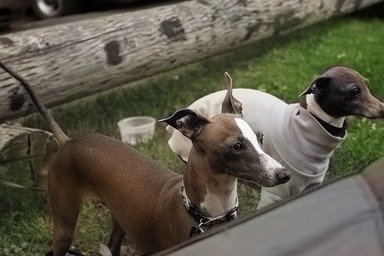}\\
            \includegraphics[width=\textwidth]{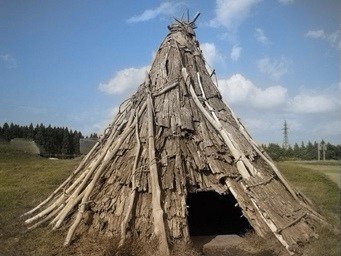}\\
            \includegraphics[width=\textwidth]{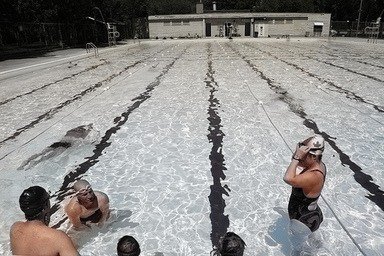}\\
            \includegraphics[width=\textwidth]{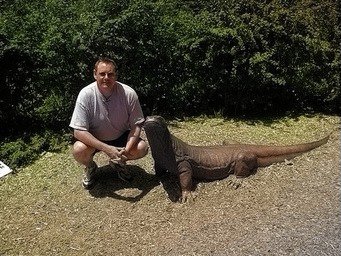}\\
        \end{center}
    \end{minipage}
    \hfill
    \begin{minipage}[b]{0.1585\linewidth}
        \begin{center}
            \includegraphics[width=\textwidth]{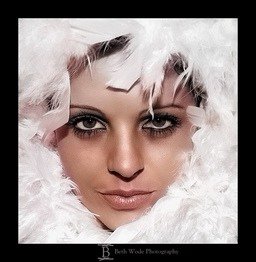}\\
            \includegraphics[width=\textwidth]{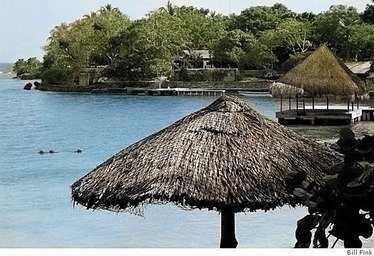}\\
            \includegraphics[width=\textwidth]{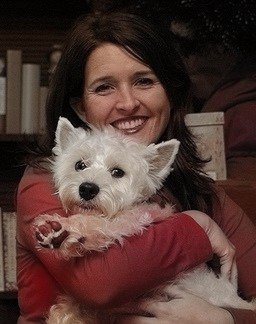}\\
            \includegraphics[width=\textwidth]{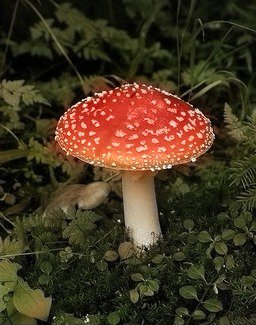}\\

        \end{center}
    \end{minipage}
    \hfill
    \begin{minipage}[b]{0.1575\linewidth}
        \begin{center}
            \includegraphics[width=\textwidth]{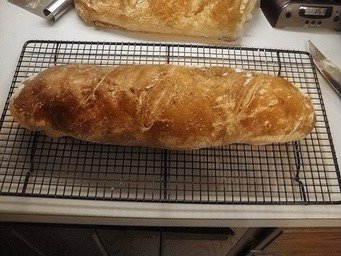}\\
            \includegraphics[width=\textwidth]{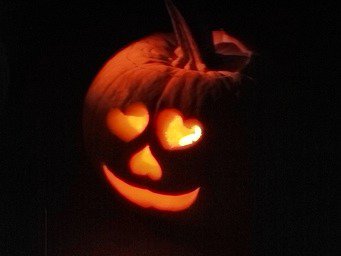}\\
            \includegraphics[width=\textwidth]{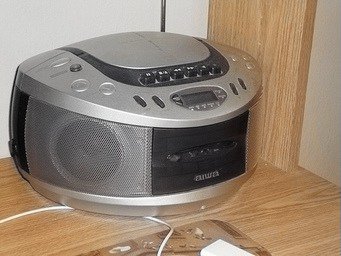}\\
            \includegraphics[width=\textwidth]{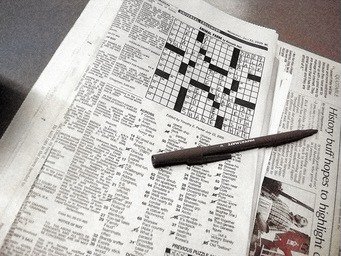}\\
            \includegraphics[width=\textwidth]{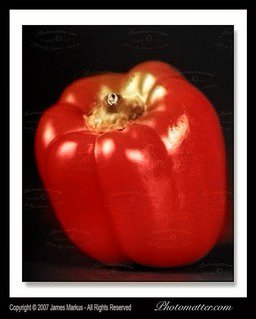}\\
        \end{center}
    \end{minipage}
    \hfill
    \begin{minipage}[b]{0.1575\linewidth}
        \begin{center}
            \includegraphics[width=\textwidth]{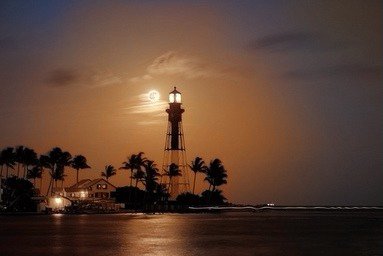}\\
            \includegraphics[width=\textwidth]{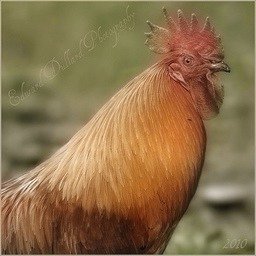}\\
            \includegraphics[width=\textwidth]{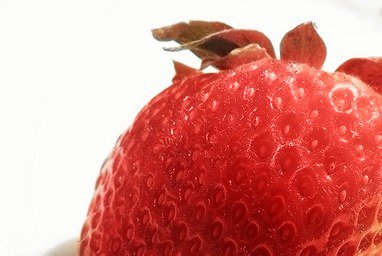}\\
            \includegraphics[width=\textwidth]{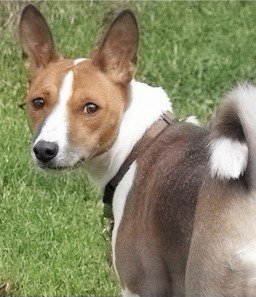}\\
            \includegraphics[width=\textwidth]{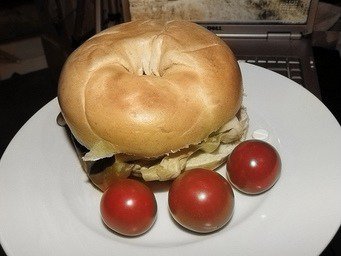}\\
        \end{center}
    \end{minipage}
    \hfill
    \begin{minipage}[b]{0.162\linewidth}
        \begin{center}
            \includegraphics[width=\textwidth]{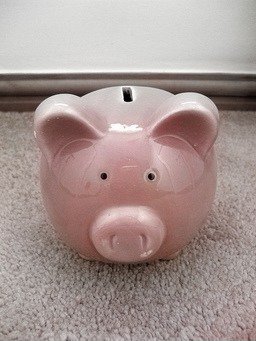}\\
            \includegraphics[width=\textwidth]{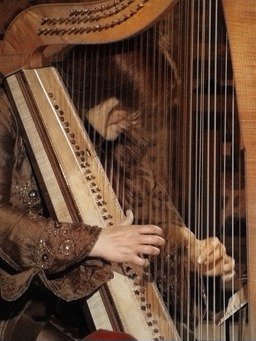}\\
            \includegraphics[width=\textwidth]{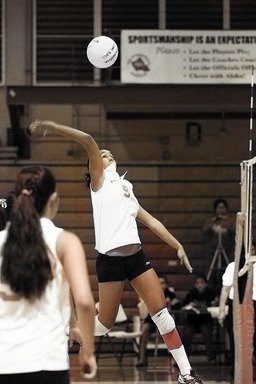}\\
        \end{center}
    \end{minipage}
    \hfill
    \begin{minipage}[b]{0.162\linewidth}
        \begin{center}
            \includegraphics[width=\textwidth]{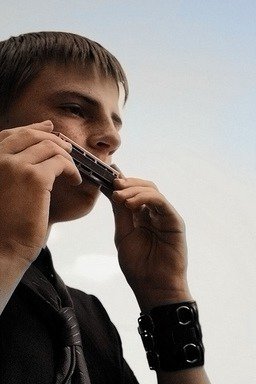}\\
            \includegraphics[width=\textwidth]{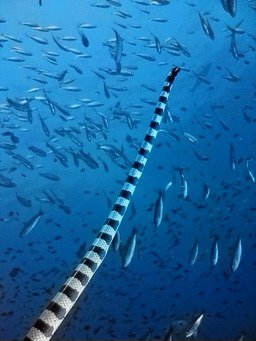}\\
            \includegraphics[width=\textwidth]{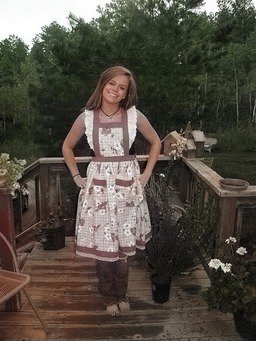}\\
        \end{center}
    \end{minipage}
    \end{center}
    \vspace{-0.05\linewidth}
    \rule{1.0\linewidth}{0.5pt}
    \vspace{-0.08\linewidth}
    \begin{center}
    \begin{minipage}[t]{0.3425\linewidth}
        \vspace{0pt}
        \begin{center}
            \includegraphics[width=.485\textwidth]{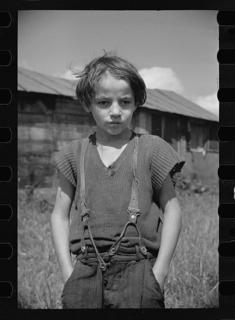}
            \includegraphics[width=.485\textwidth]{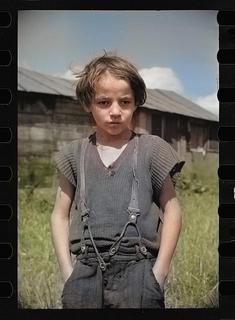}
        \end{center}
    \end{minipage}
    \hfill
    \begin{minipage}[t]{0.32\linewidth}
        \vspace{0pt}
        \begin{center}
            \includegraphics[width=.48\textwidth]{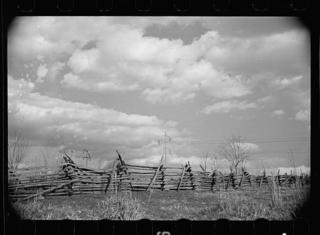}
            \includegraphics[width=.48\textwidth]{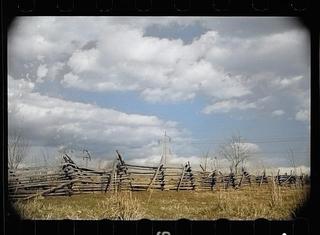}\\
            \includegraphics[width=.48\textwidth]{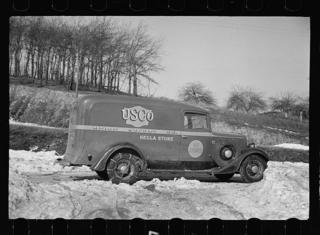}
            \includegraphics[width=.48\textwidth]{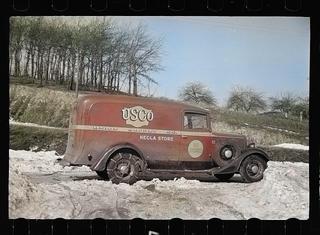}
        \end{center}
    \end{minipage}
    \hfill
    \begin{minipage}[t]{0.32\linewidth}
        \vspace{0pt}
        \begin{center}
            \includegraphics[width=.48\textwidth]{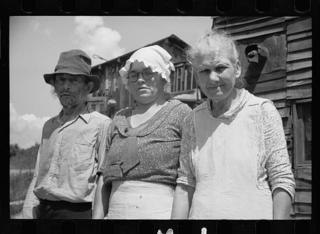}
            \includegraphics[width=.48\textwidth]{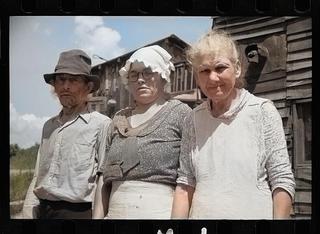}\\
            \includegraphics[width=.48\textwidth]{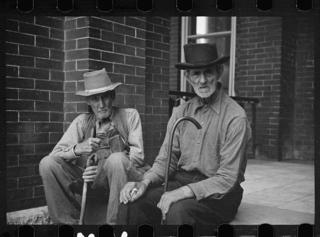}
            \includegraphics[width=.48\textwidth]{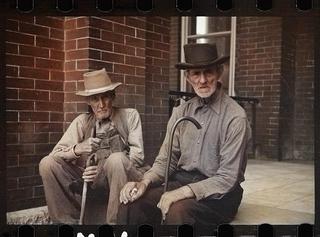}
        \end{center}
    \end{minipage}
    \end{center}
    \vspace{-0.03\linewidth}
    \caption{\small
        \textbf{Additional results.}
        \emph{Top:}
            Our automatic colorizations of these ImageNet examples
            are difficult to distinguish from real color images.
        \emph{Bottom:} B\&W photographs.
    }
    \label{fig:ctest10k-more-examples}
\end{figure}

\begin{figure}[!th]
    \RawFloats
    \setlength\fboxsep{0pt}
    \begin{center}
        \begin{minipage}[t]{0.1625\linewidth}
        \begin{center}
            \begin{minipage}[t]{0.975\linewidth}
            \begin{center}
                \includegraphics[width=\textwidth]{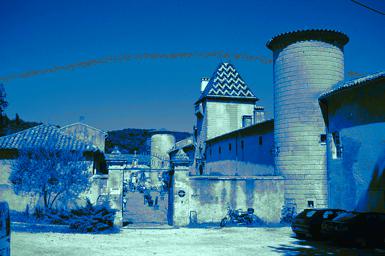}\\
                \includegraphics[width=\textwidth]{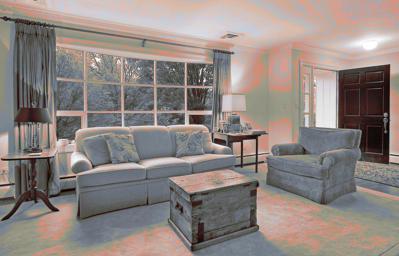}\\
                \includegraphics[width=\textwidth]{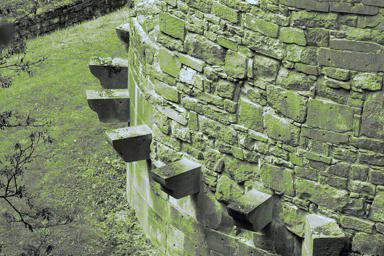}\\
                \scriptsize{\textbf{\textsf{Grayscale only}}}
            \end{center}
            \end{minipage}\\
            \rule{0.95\linewidth}{0.5pt}\\
            \scriptsize{\textbf{\textsf{Welsh~\etal~\cite{welsh2002transferring}}}}
        \end{center}
        \end{minipage}
        \hfill
        \begin{minipage}[t]{0.325\linewidth}
        \begin{center}
            \begin{minipage}[t]{0.4875\linewidth}
            \begin{center}
                \includegraphics[width=\textwidth]{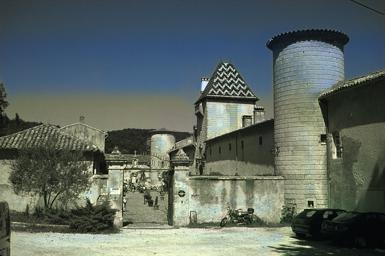}\\
                \includegraphics[width=\textwidth]{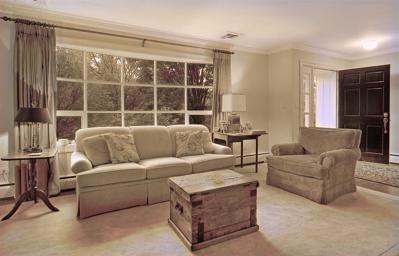}\\
                \includegraphics[width=\textwidth]{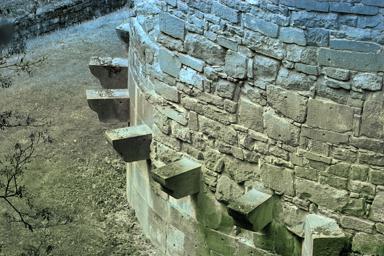}\\
                \scriptsize{\textbf{\textsf{\textcolor{white}{y}GT Scene\textcolor{white}{y}}}}
            \end{center}
            \end{minipage}
            \begin{minipage}[t]{0.4875\linewidth}
            \begin{center}
                \includegraphics[width=\textwidth]{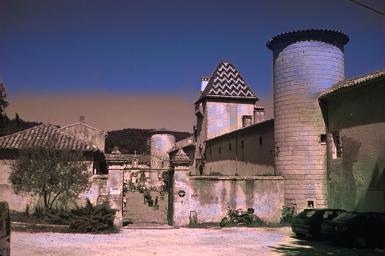}\\
                \includegraphics[width=\textwidth]{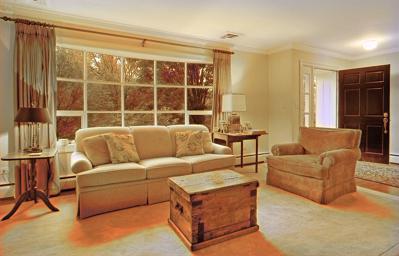}\\
                \includegraphics[width=\textwidth]{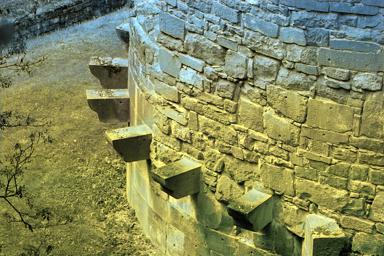}\\
                \scriptsize{\textbf{\textsf{GT Scene \& Hist}}}
            \end{center}
            \end{minipage}\\
            \rule{0.975\linewidth}{0.5pt}\\
            \scriptsize{\textbf{\textsf{Deshpande~\etal~\cite{deshpande2015learning}}}}
        \end{center}
        \end{minipage}
        \hfill
        \begin{minipage}[t]{0.325\linewidth}
        \begin{center}
            \begin{minipage}[t]{0.4875\linewidth}
            \begin{center}
                \includegraphics[width=\textwidth]{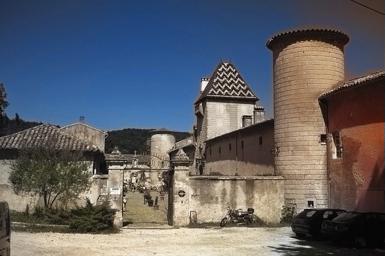}\\
                \includegraphics[width=\textwidth]{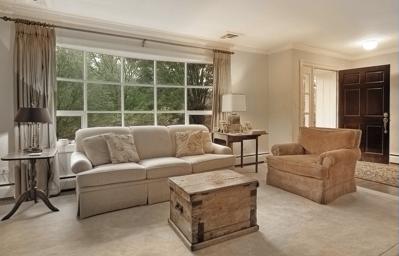}\\
                \includegraphics[width=\textwidth]{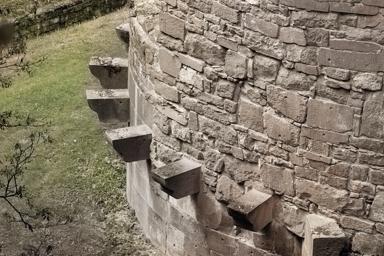}\\
                \scriptsize{\textbf{\textsf{Grayscale only}}}
            \end{center}
            \end{minipage}
            \begin{minipage}[t]{0.4875\linewidth}
            \begin{center}
                \includegraphics[width=\textwidth]{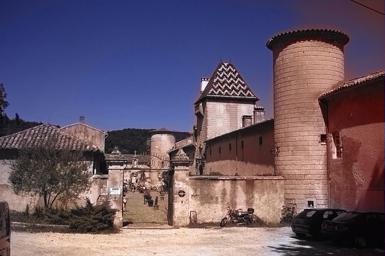}\\
                \includegraphics[width=\textwidth]{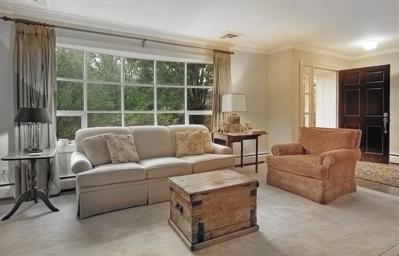}\\
                \includegraphics[width=\textwidth]{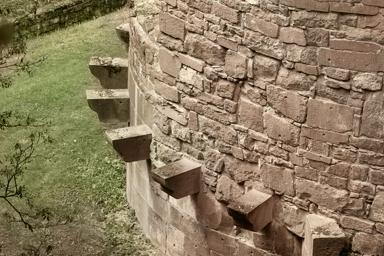}\\
                \scriptsize{\textbf{\textsf{GT Histogram}}}
            \end{center}
            \end{minipage}\\
            \rule{0.975\linewidth}{0.5pt}\\
            \scriptsize{\textbf{\textsf{Our Method}}}
        \end{center}
        \end{minipage}
        \hfill
        \begin{minipage}[t]{0.1625\linewidth}
        \begin{center}
            \begin{minipage}[t]{0.975\linewidth}
            \begin{center}
                \includegraphics[width=\textwidth]{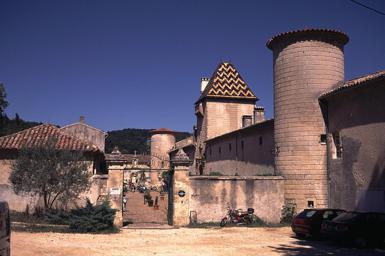}\\
                \includegraphics[width=\textwidth]{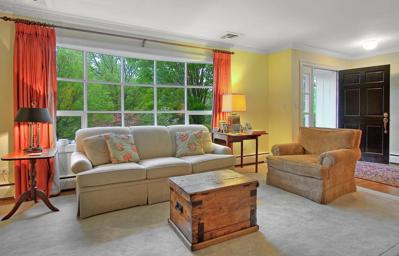}\\
                \includegraphics[width=\textwidth]{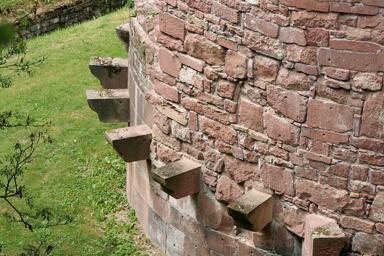}\\
                \scriptsize{\textbf{\textsf{Ground-truth}}}
            \end{center}
            \end{minipage}
        \end{center}
        \end{minipage}
    \end{center}
    \caption{\small
        \textbf{SUN-6.}
        GT Scene: test image scene class is available.
        GT Hist: test image color histogram is available.
        We obtain colorizations with visual quality better than those from
        prior work, even though we do not exploit reference images or known
        scene class.  Our energy minimization method
        (Section~\ref{sec:method_transfer}) for GT Hist further
        improves results.  In either mode, our method appears less dependent
        on spatial priors: note splitting of the sky in the first row and
        correlation of green with actual grass in the last row.
    }
    \label{fig:sun6-examples}
\end{figure}

\section{Experiments}
\label{sec:experiments}

Starting from pretrained VGG-16-Gray, described in the previous
section, we attach \texttt{h\_fc1} and output prediction layers with
Xavier initialization~\cite{glorot2010understanding}, and fine-tune the
entire system for colorization.  We consider multiple prediction layer
variants: Lab output with $L_2$ loss, and both Lab and hue/chroma marginal or
joint histogram output with losses according to
Equations~\eqref{eq:loss-hist} and~\eqref{eq:loss-hc}.  We train each system
variant end-to-end for one epoch on the $1.2$ million images of the ImageNet
training set, each resized to at most $256$ pixels in smaller dimension.
A single epoch takes approximately 17 hours on a GTX Titan X GPU. At test
time, colorizing a single $512 \times 512$ pixel image takes $0.5$ seconds.

\begin{table}[t!]
\RawFloats
\begin{minipage}[t]{.52\textwidth}
\centering
    {\small
    \begin{tabular*}{\textwidth}{l @{\extracolsep{\fill}} rr}
        \toprule
        Model$\backslash$Metric  & RMSE        & PSNR      \\
        \midrule
        No colorization          & 0.343       & 22.98    \\ % c53
        Lab, $L_2$               & 0.318       & 24.25    \\ % c53
        Lab, $K = 32$            & 0.321       & 24.33    \\ % c55
        Lab, $K = 16 \times 16$  & 0.328       & 24.30    \\ % c57
        Hue/chroma, $K = 32$     & 0.342       & 23.77    \\ % c56
        \quad + chromatic fading & {\bf 0.299} & {\bf 24.45}  \\
        \bottomrule
    \end{tabular*}}
    \caption{
        \small
        \textbf{ImageNet/cval1k.}
        Validation performance of system variants.
        Hue/chroma is best, but only with chromatic fading.
    }
    \label{tab:cval1k}
\end{minipage}
\hfill
\begin{minipage}[t]{.45\textwidth}
\centering
    {\small
    \begin{tabular*}{\textwidth}{l @{\extracolsep{\fill}} rr}
        \toprule
        Model$\backslash$Metric & RMSE        & PSNR    \\
        \midrule
        \texttt{data..fc7}      & {\bf 0.299} & {\bf 24.45} \\ % c56
        \texttt{data..conv5\_3} & 0.306       & 24.13   \\ % c63
        \texttt{conv4\_1..fc7}  & 0.302       & {\bf 24.45} \\ % c64
        \texttt{conv5\_1..fc7}  & 0.307       & 24.38        \\ % c68
        \texttt{fc6..fc7}       & 0.323       & 24.22   \\ % c66
        \texttt{fc7}            & 0.324       & 24.19   \\ % c65
        \bottomrule
    \end{tabular*}}
    \caption{
        \small
        \textbf{ImageNet/cval1k.}
        Ablation study of hypercolumn components.
    }
    \label{tab:ablation}
\end{minipage}
\end{table}

We setup two disjoint subsets of the ImageNet validation data for our own use:
$1000$ validation images (\textbf{cval1k}) and $10000$ test images
(\textbf{ctest10k}).  Each set has a balanced representation for ImageNet
categories, and excludes any images encoded as grayscale, but may include
images that are naturally grayscale (\eg~closeup of nuts and bolts), where
an algorithm should know not to add color.  Category labels are
discarded; only images are available at test time.  We propose
\textbf{ctest10k} as a standard benchmark with the following metrics:
\begin{itemize}
    \item{
        \textbf{RMSE}:
        root mean square error in $\alpha\beta$ averaged over all
        pixels~\cite{deshpande2015learning}.
    }
    \item{
        \textbf{PSNR}:
        peak signal-to-noise ratio in RGB calculated per
        image~\cite{cheng2015deep}.  We use the arithmetic mean of PSNR over
        images, instead of the geometric mean as in
        Cheng~\etal~\cite{cheng2015deep};
        geometric mean is overly sensitive to outliers.
    }
\end{itemize}
By virtue of comparing to ground-truth color images, quantitative
colorization metrics can penalize reasonable, but incorrect, color guesses for
many objects (\eg~red car instead of blue car) more than jarring artifacts.
This makes qualitative results for colorization as important as quantitative;
we report both.

Figures~\ref{fig:teaser},~\ref{fig:ctest10k-examples}, and~\ref{fig:ctest10k-more-examples}
show example test results of our best system variant, selected according to
performance on the validation set and trained for a total of $10$ epochs.  This
variant predicts hue and chroma and
uses chromatic fading during image generation.  Table~\ref{tab:cval1k} provides
validation benchmarks for all system variants, including the trivial baseline
of no colorization.  On ImageNet test (\textbf{ctest10k}), our selected model
obtains $0.293$ (RMSE, $\alpha\beta$, avg/px) and $24.94$~dB (PSNR, RGB,
avg/im), compared to $0.333$ and $23.27$~dB for the baseline.

Table~\ref{tab:ablation} examines the importance of different neural network
layers to colorization; it reports validation performance of ablated systems
that include only the specified subsets of layers in the hypercolumn used to
predict hue and chroma.  Some lower layers may be discarded without much
performance loss, yet higher layers alone (\texttt{fc6..fc7}) are insufficient
for good colorization.

Our ImageNet colorization benchmark is new to a field lacking an
established evaluation protocol.  We therefore focus on comparisons with
two recent papers~\cite{deshpande2015learning,cheng2015deep}, using their
self-defined evaluation criteria.  To do so, we run our ImageNet-trained
hue and chroma model on two additional datasets:
\begin{itemize}
    \item{
        \textbf{\bf SUN-A}~\cite{patterson2014sun} is a subset of the SUN
        dataset~\cite{xiao2010sun} containing $47$ object categories.
        Cheng~\etal~\cite{cheng2015deep} train a colorization system on
        $2688$ images and report results on $1344$ test images.  We were
        unable to obtain the list of test images, and therefore report results
        averaged over five random subsets of $1344$ SUN-A images.  We do
        not use any SUN-A images for training.
    }
    \item{
        \textbf{SUN-6}, another SUN subset, used by
        Deshpande~\etal~\cite{deshpande2015learning}, includes images
        from $6$ scene categories (beach, castle, outdoor, kitchen,
        living room, bedroom).  We compare our results on $240$ test images
        to those reported in~\cite{deshpande2015learning} for their method
        as well as for Welsh~\etal~\cite{welsh2002transferring} with
        automatically matched reference images as
        in~\cite{morimoto2009automatic}.
        Following~\cite{deshpande2015learning}, we consider another evaluation
        regime in which ground-truth target color histograms are available.
    }
\end{itemize}
\begin{table}[!t]
\RawFloats
\begin{minipage}[t]{.40\textwidth}
    {\small
    \begin{tabular*}{\textwidth}[t]{l @{\extracolsep{\fill}} r}
        \toprule
        Method                                       & RMSE           \\
        \midrule
        Grayscale (no colorization)                  & 0.285          \\
        Welsh~\etal~\cite{welsh2002transferring}     & 0.353          \\
        Deshpande~\etal~\cite{deshpande2015learning} & 0.262          \\
        \quad + GT Scene                             & 0.254          \\
        Our Method                                   & \textbf{0.211} \\
        \bottomrule
    \end{tabular*}}
    \caption{
        \small
        \textbf{SUN-6.}
        Comparison with competing methods.
    }
    \label{tab:sun6}
\end{minipage}
\hfill
\begin{minipage}[t]{.56\textwidth}
    {\small
    \begin{tabular*}{\textwidth}[t]{l @{\extracolsep{\fill}} r}
        \toprule
        Method                                           & RMSE \\
        \midrule
        Deshpande~\etal~(C)~\cite{deshpande2015learning} & 0.236 \\
        Deshpande~\etal~(Q)                              & 0.211 \\
        Our Method (Q)                                   & 0.178 \\
        Our Method (E)                                   & \textbf{0.165} \\
        \bottomrule
    \end{tabular*}}
    \caption{
        \small
        \textbf{SUN-6 (GT Hist).}
        Comparison using ground-truth histograms.
        Results for Deshpande~\etal~\cite{deshpande2015learning} use GT Scene.
    }
    \label{tab:sun6-gth}
\end{minipage}
\end{table}
\begin{figure}[!t]
    \RawFloats
    \begin{minipage}[c]{.45\linewidth}
        \centering
        \includegraphics[width=.78\linewidth]{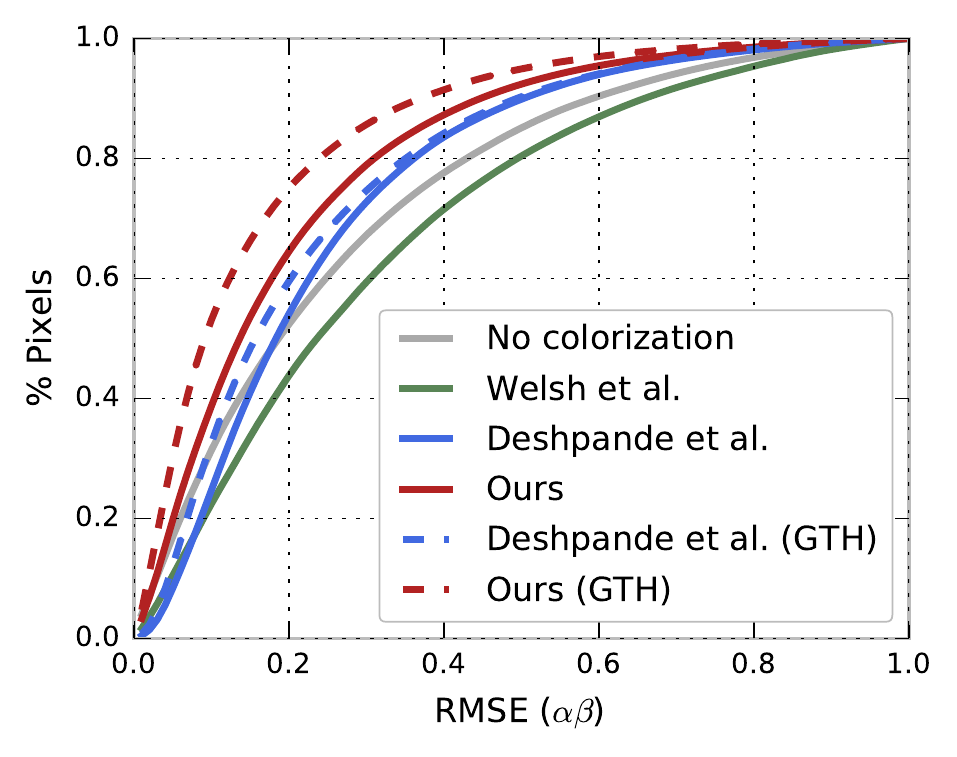}\\
        \caption{
            \small
            \textbf{SUN-6.}
            Cumulative histogram of per pixel error
            (higher=more pixels with lower error).
            Results for Deshpande~\etal~\cite{deshpande2015learning}
            use GT Scene.
        }
        \label{fig:sun6-cum-hist}
    \end{minipage}\hspace{1em}%
    \begin{minipage}[c]{.5\linewidth}
        \centering
        \includegraphics[width=.82\linewidth]{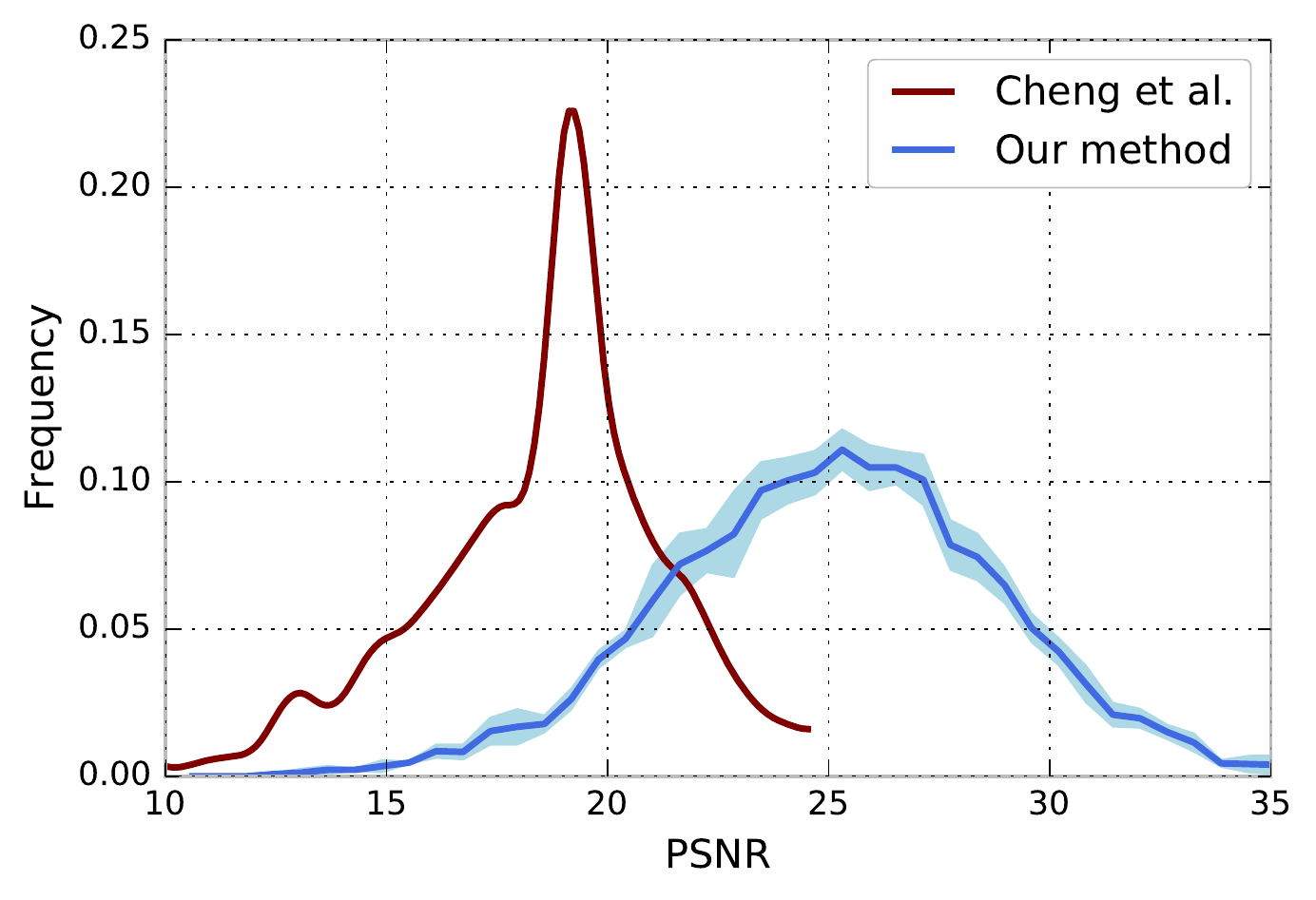}
        \caption{
            \small
            \textbf{SUN-A.}
            Histogram of per-image PSNR for~\cite{cheng2015deep} and our
            method. The highest geometric mean PSNR reported for experiments
            in~\cite{cheng2015deep} is 24.2, vs. {\bf 32.7$\pm$2.0} for us.
        }
        \label{fig:sun-deep-histogram}
\end{minipage}
\end{figure}
Figure~\ref{fig:sun6-examples} shows a comparison of results on SUN-6.
Forgoing usage of ground-truth global histograms, our fully automatic
system produces output qualitatively superior to methods relying on
such side information.  Tables~\ref{tab:sun6}~and~\ref{tab:sun6-gth}
report quantitative performance corroborating this view.  The
partially automatic systems in Table~\ref{tab:sun6-gth} adapt output
to fit global histograms using either: (C) cluster
correspondences~\cite{deshpande2015learning}, (Q) quantile matching, or
(E) our energy minimization described in Section~\ref{sec:method_transfer}.
Our quantile matching results are superior to those
of~\cite{deshpande2015learning} and our new energy minimization procedure
offers further improvement.

Figures~\ref{fig:sun6-cum-hist} and~\ref{fig:sun-deep-histogram} compare
error distributions on SUN-6 and SUN-A.  As in Table~\ref{tab:sun6}, our
fully automatic method dominates all competing approaches, even those
which use auxiliary information.  It is only outperformed by the version of
itself augmented with ground-truth global histograms.  On SUN-A,
Figure~\ref{fig:sun-deep-histogram} shows clear separation between our
method and~\cite{cheng2015deep} on per-image PSNR.

The Appendix (Figures~\ref{fig:charpiat-reference} and~\ref{fig:charpiat-portraits}) provides anecdotal comparisons to one additional
method, that of Charpiat~\etal~\cite{charpiat2010machine}, which can be
considered an automatic system if reference images are available.
Unfortunately, source code of~\cite{charpiat2010machine} is not
available and reported time cost is prohibitive for large-scale
evaluation ($30$ minutes per image).  We were thus unable to
benchmark~\cite{charpiat2010machine} on large datasets.

With regard to concurrent work, Zhang~\etal~\cite{zhang2016colorful} include a
comparison of our results to their own.  The two systems are competitive in
terms of quantitative measures of colorization accuracy.  Their system, set to
produce more vibrant colors, has an advantage in terms of human-measured
preferences.  In contrast, an off-the-shelf VGG-16 network for image
classification, consuming our system's color output, more often produces
correct labels, suggesting a realism advantage.  We refer interested readers
to~\cite{zhang2016colorful} for the full details of this comparison.

\begin{figure}[!t]
    \begin{center}
    \begin{minipage}[b]{0.1880\linewidth}
        \vspace{0pt}
        \begin{center}
            \includegraphics[width=\textwidth]{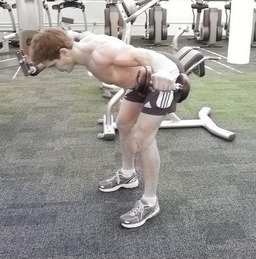}\\
            \includegraphics[width=\textwidth]{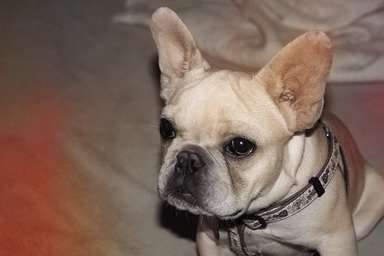}
        \end{center}
    \end{minipage}
    \hfill
    \begin{minipage}[b]{0.2045\linewidth}
        \vspace{0pt}
        \begin{center}
            \includegraphics[width=\textwidth]{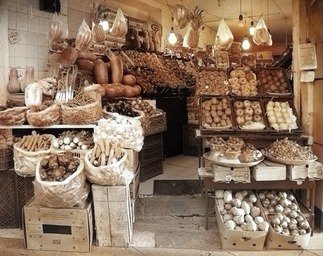}\\
            \includegraphics[width=\textwidth]{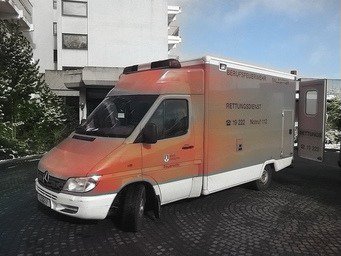}
        \end{center}
    \end{minipage}
    \hfill
    \begin{minipage}[b]{0.1802\linewidth}
        \vspace{0pt}
        \begin{center}
            \includegraphics[width=\textwidth]{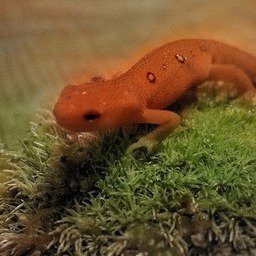}\\
            \includegraphics[width=\textwidth]{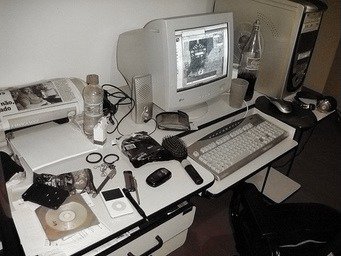}
        \end{center}
    \end{minipage}
    \hfill
    \begin{minipage}[b]{0.1895\linewidth}
        \vspace{0pt}
        \begin{center}
            \includegraphics[width=\textwidth]{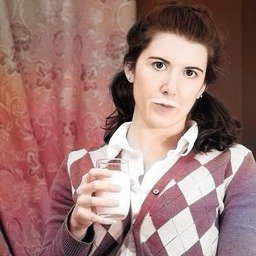}\\
            \includegraphics[width=\textwidth]{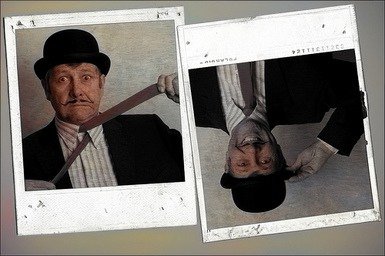}
        \end{center}
    \end{minipage}
    \hfill
    \begin{minipage}[b]{0.1860\linewidth}
        \vspace{0pt}
        \begin{center}
            \includegraphics[width=\textwidth]{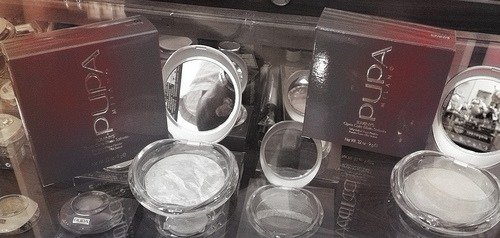}\\
            \includegraphics[width=\textwidth]{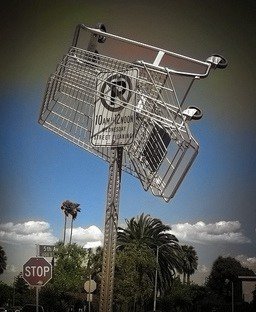}
        \end{center}
    \end{minipage}
    \vspace{-0.25cm}
    \end{center}
    \caption{\small
        \textbf{Failure modes.}
        \emph{Top row, left-to-right:}
            texture confusion,
            too homogeneous,
            color bleeding,
            unnatural color shifts ($\times 2$).
        \emph{Bottom row:}
            inconsistent background,
            inconsistent chromaticity,
            not enough color,
            object not recognized (upside down face partly gray),
            context confusion (sky).
    }
    \label{fig:failure-modes}
\end{figure}

Though we achieve significant improvements over prior state-of-the-art, our
results are not perfect.  Figure~\ref{fig:failure-modes} shows examples of
significant failures.  Minor imperfections are also present in some of the
results in Figures~\ref{fig:ctest10k-examples}
and~\ref{fig:ctest10k-more-examples}.  We believe a common failure mode
correlates with gaps in semantic interpretation: incorrectly identified or
unfamiliar objects and incorrect segmentation.  In addition, there are
``mistakes'' due to natural uncertainty of color -- \eg~the graduation robe
at the bottom right of Figure~\ref{fig:ctest10k-examples} is red, but could
as well be purple.

Since our method produces histograms, we can provide interactive means of
biasing colorizations according to user preferences.  Rather than output a
single color per pixel, we can sample color for image regions and evaluate
color uncertainty.  Specifically, solving our energy minimization formulation
(Equation~\eqref{eq:energy-min}) with global biases~$\mathbf{b}$ that are not
optimized based on a reference image, but simply ``rotated'' through color
space, induces changed color preferences throughout the image.  The uncertainty
in the predicted histogram modulates this effect.

\begin{figure}[!t]
    \RawFloats
    \centering
    \begin{tabular}{ccccc}
        \includegraphics[width=.190\textwidth]{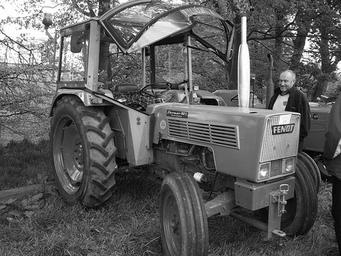}&
        \includegraphics[width=.190\textwidth]{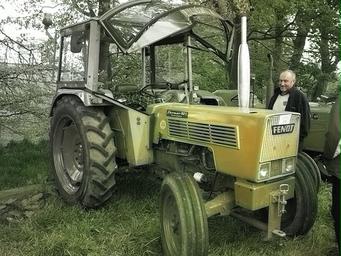}&
        \includegraphics[width=.190\textwidth]{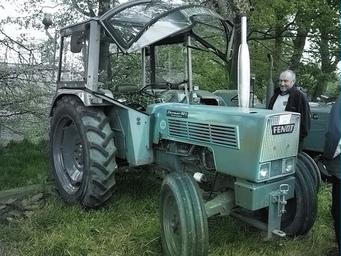}&
        \includegraphics[width=.190\textwidth]{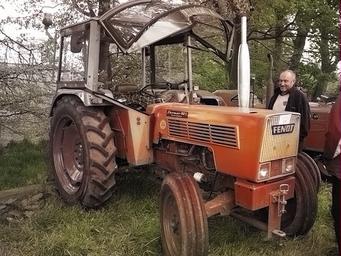}&
        \includegraphics[width=.190\textwidth]{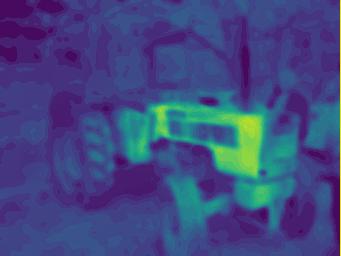}
    \end{tabular}
    \caption{
        \small
        \textbf{Sampling colorizations.}
        \emph{Left:} Image \& 3 samples;
        \emph{Right:} Uncertainty map.
    }
    \label{fig:warhol}
\end{figure}

Figure~\ref{fig:warhol} shows multiple sampled colorizations, together with a
visualization of uncertainty.  Here, uncertainty is the entropy of the
predicted hue multiplied by the chroma.  Our distributional output and energy
minimization framework open the path for future investigation of
human-in-the-loop colorization tools.

\subsection{Representation learning}
\label{sec:representation-learning}

High-level visual understanding is essential for the colorization of grayscale
images, motivating our use of an ImageNet pretrained network as a starting
point.  But with enough training data, perhaps we can turn this around and
use colorization as means of learning networks for capturing high-level visual
representations.  Table~\ref{tab:from-scratch} shows that a colorization
network, trained from scratch using only unlabeled color images, is
surprisingly competitive.  It converges slower, but requires not more than
twice the number of epochs.

Our preliminary work shows that the networks learned via training colorization
from scratch generalize well to other visual tasks. This is significant because
such training requires no human annotation effort.  It follows a recent trend
of learning representations through self-supervision (\eg{} context
prediction~\cite{contextpred}, solving jigsaw puzzles~\cite{jigsaw},
inpainting~\cite{contextencoders}, adversarial feature learning \cite{bigan,dumoulin2016adversarially}).

We examine self-supervised colorization as a replacement for supervised
ImageNet pretraining on the Pascal VOC 2012 semantic segmentation task, with
results on grayscale validation set images.  We train colorization from
scratch on ImageNet (Table~\ref{tab:from-scratch}) and fine-tune for Pascal
semantic segmentation.  We make the one adjustment of employing cross-validated
early stopping to avoid overfitting.  Table~\ref{tab:voc2012-segmentation}
shows this strategy to be promising as a drop-in replacement for supervised
ImageNet pretraining.  Self-supervised colorization more than halfway bridges
the gap between random initialization and supervised pretraining.

As VGG-16 is a more performant architecture, comparison with prior work
is not straightforward. Yet, Table~\ref{tab:voc2012-segmentation} still
indicates that colorization is a front-runner among the self-supervision
methods, leading to an 18-point improvement in mIU over the baseline. To our
knowledge, 50.2\% is the highest reported result that does not supplement
training with additional annotated data~\cite{ion2014probabilistic}.

\begin{table}[!t]
\RawFloats
\begin{minipage}[t]{.36\textwidth}
\centering
    {\small
    \begin{tabular*}{\textwidth}[t]{l @{\extracolsep{\fill}} rr}
        \toprule
        Initialization            & RMSE   &  PSNR  \\
        \midrule
        Classifier                & 0.299  &  24.45 \\
        Random                    & 0.311  &  24.25 \\
        \bottomrule
    \end{tabular*}}
    \caption{
        \small
        \textbf{ImageNet/cval1k.}
        Compares methods of initialization before colorization training. Hue/chroma with chromatic fading is used in both cases (see in Tab.~\ref{tab:cval1k}).
    }
    \label{tab:from-scratch}
\end{minipage}
\hfill
\begin{minipage}[t]{.60\textwidth}
\centering
    {\small
    \begin{tabular*}{\textwidth}[t]{l @{\extracolsep{\fill}} l @{\extracolsep{\fill}} c @{\extracolsep{\fill}} c @{\extracolsep{\fill}} c @{\extracolsep{\fill}} r}
        \toprule
        Initialization                                 & Architecture & $X$    & $Y$    & $C$    & mIU (\%) \\ \midrule
        Classifier                                     & VGG-16  & \cmark & \cmark &        & 64.0 \\ %\midrule
        Colorizer                                      & VGG-16  & \cmark &        &        & 50.2 \\
        Random                                         & VGG-16  &        &        &        & 32.5 \\ \midrule
        Classifier~\cite{bigan,contextencoders}        & AlexNet & \cmark & \cmark & \cmark & 48.0 \\
        BiGAN~\cite{bigan}                             & AlexNet & \cmark &        & \cmark & 34.9 \\
        Inpainter~\cite{contextencoders}               & AlexNet & \cmark &        & \cmark & 29.7 \\
        Random~\cite{contextencoders}                  & AlexNet &        &        & \cmark & 19.8 \\
        \bottomrule
    \end{tabular*}}
    \caption{
        \small
        \textbf{VOC 2012 segmentation validation set.}
        %The ``Colorizer'' is the same as ``Random'' in Tab.~\ref{tab:from-scratch}.
        Pretraining uses ImageNet images ($X$), labels ($Y$). VOC 2012 images are in color ($C$).
    }
    \label{tab:voc2012-segmentation}
\end{minipage}
\end{table}

%auto-ignore
\section{Conclusion}
\label{sec:final}

We present a system that demonstrates state-of-the-art ability to automatically
colorize grayscale images.  Two novel contributions enable this progress: a
deep neural architecture that is trained end-to-end to incorporate semantically
meaningful features of varying complexity into colorization, and a color
histogram prediction framework that handles uncertainty and ambiguities
inherent in colorization while preventing jarring artifacts.  Our fully
automatic colorizer produces strong results, improving upon previously leading
methods by large margins on all datasets tested; we also propose a new
large-scale benchmark for automatic image colorization, and establish a strong
baseline with our method to facilitate future comparisons.  Our colorization
results are visually appealing even on complex scenes, and allow for effective
post-processing with creative control via color histogram transfer and
intelligent, uncertainty-driven color sampling.  We further reveal
colorization as a promising avenue for self-supervised visual learning.

~\\
\noindent
%\begin{small}
\textbf{Acknowledgements.}
We thank Ayan Chakrabarti for suggesting lightness-normalized quantile matching
and for useful discussions, and Aditya Deshpande and Jason Rock for discussions
on their work. We gratefully acknowledge the support of NVIDIA Corporation with
the donation of GPUs for this research.
%\end{small}

%\placeholder{Discussion with Ayan. GPU from NVIDIA. Thanks Aditya Deshpande.}

%\clearpage
\bibliographystyle{splncs03}
\bibliography{egbib}

%\begin{appendices}

\newpage
\appendix
%auto-ignore
\begin{figure}[h!]
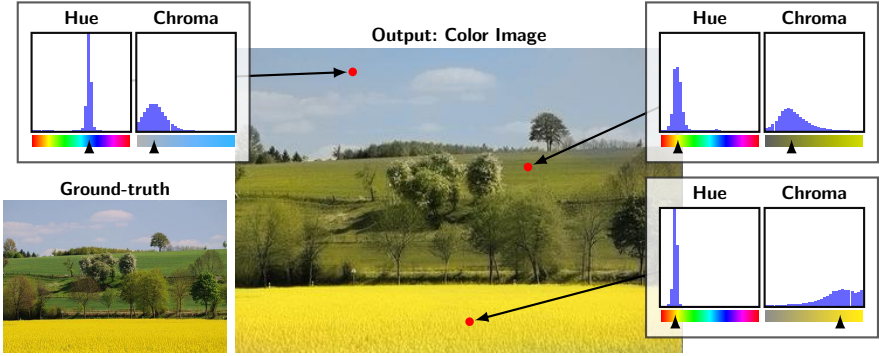

    \centering
    \includestandalone{figures/field}
    \caption{\textbf{Histogram predictions.} Example of predicted hue/chroma histograms.}
    \label{fig:histogram-supp-field}
\end{figure}

\noindent
Appendix~\ref{sec:supp-details} provides additional training and evaluation details. This is followed by more results and examples in Appendix~\ref{sec:supp-results}.

\section{Supplementary details}
\label{sec:supp-details}

\subsection{Re-balancing}
\label{sec:rebalance}

To adjust the scale of the activations of layer $l$ by factor $m$, without changing any other layer's activation, the weights $\bW$ and the bias $\bb$ are updated according to:
\begin{equation}
    \bW_l \leftarrow m \bW_l \quad\quad\quad\quad
    \bb_l \leftarrow m \bb_l \quad\quad \quad\quad
    \bW_{l+1} \leftarrow \frac 1m \bW_{l+1}
\end{equation}
The activation of $\bx_{l+1}$ becomes:
\begin{equation}
    \bx_{l+1} = \frac 1m \bW_{l+1} \mathrm{ReLU}(m \bW_l \bx_l + m \bb_l) + \bb_{l+1}
\end{equation}
The $m$ inside the ReLU will not affect whether or not a value is rectified, so the two cases remain the same: (1) negative: the activation will be the corresponding feature in $\bb_{l+1}$ regardless of $m$, and (2) positive: the ReLU becomes the identity function and $m$ and $\frac 1m$ cancel to get back the original activation.

We set $m = \frac{1}{\sqrt{\hat{\mathbb{E}}[X^2]}}$, estimated for each layer separately. 

\subsection{Color space $\alpha\beta$}
The color channels $\alpha\beta$ (``ab'' in~\cite{deshpande2015learning}) are calculated as
\begin{equation}
    \alpha = \frac{B - \frac 12 (R + G) }{L + \epsilon} \quad\quad
    \beta  = \frac{R - G}{L + \epsilon} \\
\end{equation}
where $\epsilon = 0.0001$, $R, G, B \in [0, 1]$ and $L = \frac{R+G+B}3$.\footnote{We know that this is how Deshpande~\etal~\cite{deshpande2015learning} calculate it based on their code release.}

\subsection{Error metrics}
For $M$ images, each image $m$ with $N_m$ pixels, we calculate the error metrics as:

\begin{align}
                                                                            \mathrm{RMSE} &= \frac 1{\sum_{m=1}^M N_m} \sum_{m=1}^M \sum_{n=1}^{N_m} \sqrt{\left\| \left[\by^{(m)}_{\alpha\beta}\right]_n - \left[\hat\by^{(m)}_{\alpha\beta}\right]_n \right\|^2 } \\
    \mathrm{PSNR} &= \frac 1M \sum_{m=1}^M \sum_{n=1}^{N_m} -10 \cdot \log_{10} \left(\frac{\|\by_\mathrm{RGB}^{(m)} - \hat \by_\mathrm{RGB}^{(m)}\|^2}{3N_m}\right)
\end{align}
Where $\by_{\alpha\beta}^{(m)} \in [-3, 3]^{N_m \times 2}$ and $\by_\mathrm{RGB}^{(m)} \in [0, 1]^{N_m \times 3}$ for all $m$.

\begin{figure}[t!]
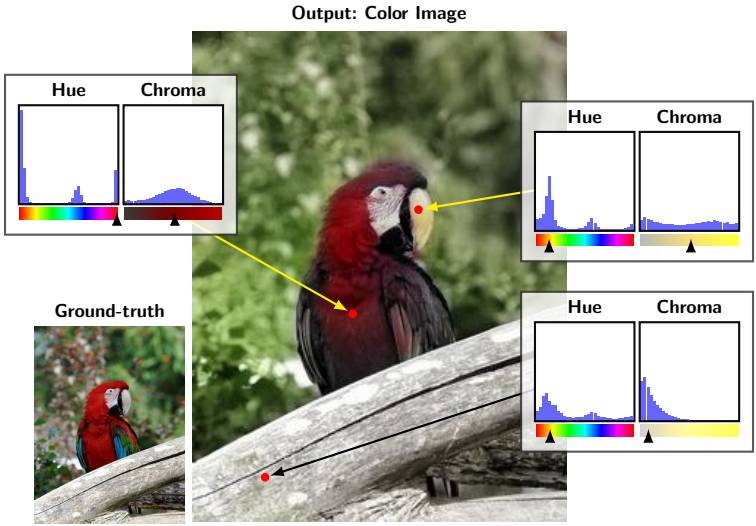

    \centering
    \includestandalone{figures/parrot}
    \caption{\textbf{Histogram predictions.} Example of predicted hue/chroma histograms.}
    \label{fig:histogram-supp-parrot}
\end{figure}

\subsection{Lightness correction}
Ideally the lightness $L$ is an unaltered pass-through channel. However, due to
subtle differences in how $L$ is defined, it is possible that the lightness of
the predicted image, $\hat L$, does not agree with the input, $L$. To
compensate for this, we add $L - \hat L$ to all color channels in the
predicted RGB image as a final corrective step.

%\newpage
\section{Supplementary results}
\label{sec:supp-results}

\subsection{Validation}
A more detailed list of validation results for hue/chroma inference methods is seen in Table~\ref{tab:supp-cval1k-hue-chroma}.

\begin{table}[t!]
    \RawFloats
    \floatbox[{\capbeside\thisfloatsetup{
        capbesideposition={right,top},capbesidewidth=5cm}}]{figure}[\FBwidth]
    {
        \captionof{table}{
            \small
            \textbf{ImageNet/cval1k.}
            Comparison of various histogram inference methods for hue/chroma.
            Mode/mode does fairly well but has severe visual artifacts.
            (CF = Chromatic fading)
        }
        \label{tab:supp-cval1k-hue-chroma}
    }
    {
        {\small
        \begin{tabular*}{0.52\textwidth}{llc @{\extracolsep{\fill}} rr}
            \toprule
            Hue         & Chroma      & CF  & RMSE        & PSNR        \\ 
            \midrule
            Sample      & Sample      &     & 0.426       & 21.41       \\
            Mode        & Mode        &     & 0.304       & 23.90       \\
            Expectation & Expectation &     & 0.374       & 23.13       \\
            Expectation & Expectation & \cmark & 0.307       & 24.35       \\
            Expectation & Median      &     & 0.342       & 23.77       \\
            Expectation & Median      & \cmark & {\bf 0.299} & {\bf 24.45} \\
            \bottomrule
        \end{tabular*}}
    }
\end{table}

\subsection{Examples}

\begin{figure}[!ht]
    \RawFloats
    \centering
    \begin{tabular}{ccccc}
        \includegraphics[width=.190\textwidth]{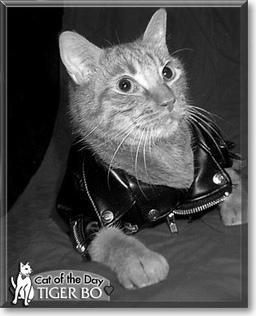}&
        \includegraphics[width=.190\textwidth]{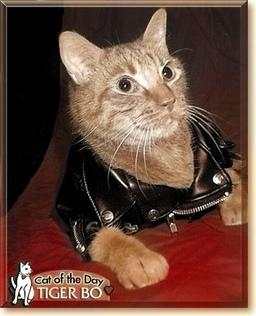}&
        \includegraphics[width=.190\textwidth]{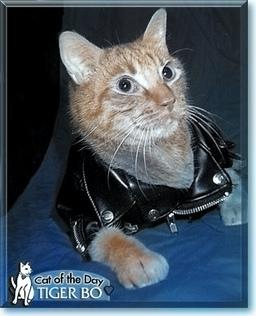}&
        \includegraphics[width=.190\textwidth]{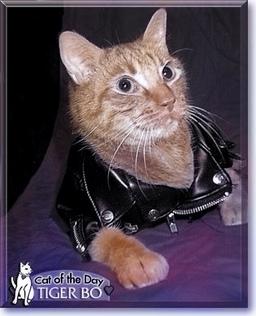}&
        \includegraphics[width=.190\textwidth]{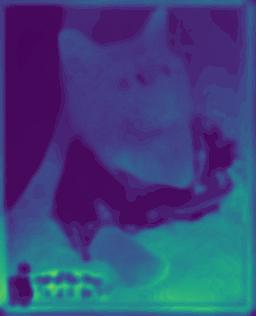}\\
        \includegraphics[width=.190\textwidth]{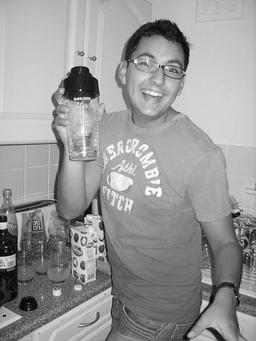}&
        \includegraphics[width=.190\textwidth]{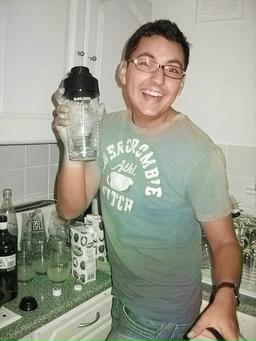}&
        \includegraphics[width=.190\textwidth]{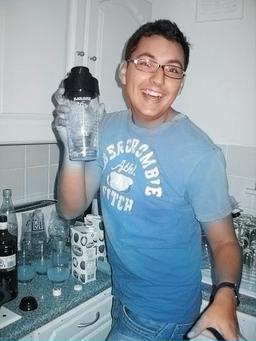}&
        \includegraphics[width=.190\textwidth]{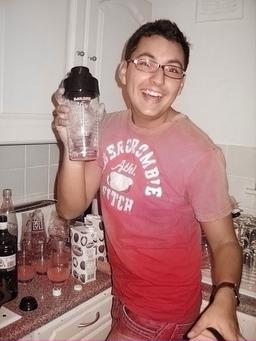}&
        \includegraphics[width=.190\textwidth]{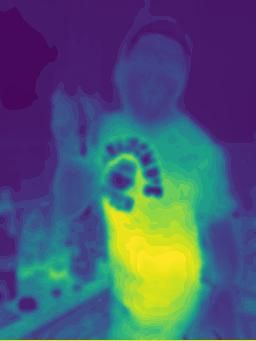}\\
        \includegraphics[width=.190\textwidth]{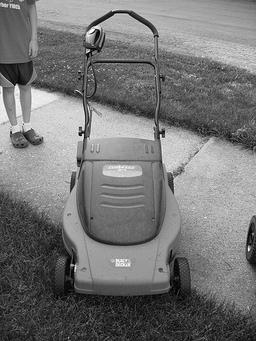}&
        \includegraphics[width=.190\textwidth]{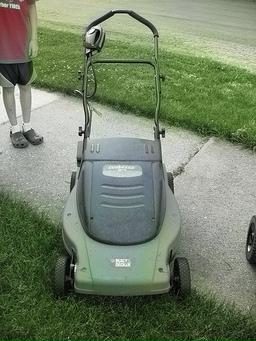}&
        \includegraphics[width=.190\textwidth]{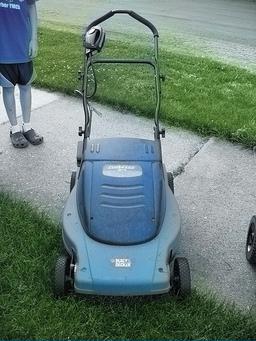}&
        \includegraphics[width=.190\textwidth]{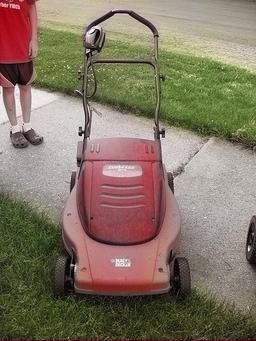}&
        \includegraphics[width=.190\textwidth]{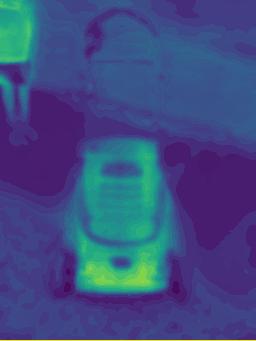}\\
        \includegraphics[width=.190\textwidth]{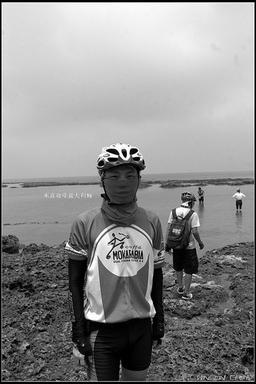}&
        \includegraphics[width=.190\textwidth]{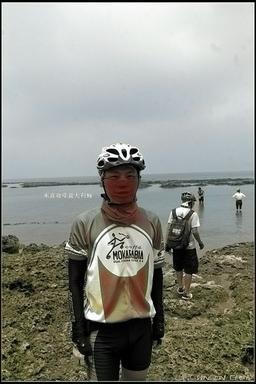}&
        \includegraphics[width=.190\textwidth]{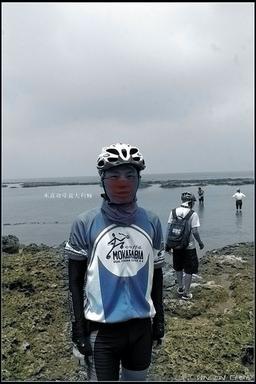}&
        \includegraphics[width=.190\textwidth]{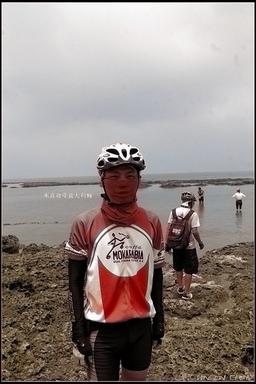}&
        \includegraphics[width=.190\textwidth]{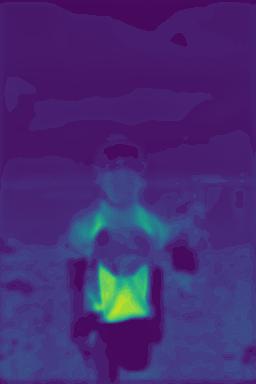}\\
        \includegraphics[width=.190\textwidth]{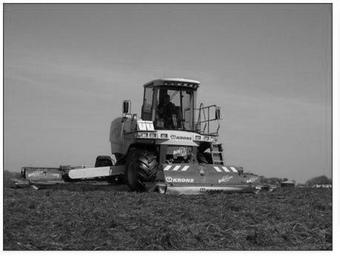}&
        \includegraphics[width=.190\textwidth]{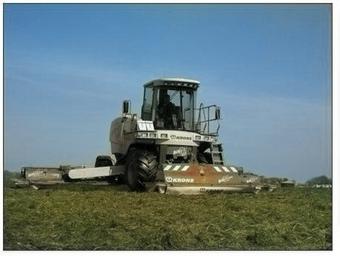}&
        \includegraphics[width=.190\textwidth]{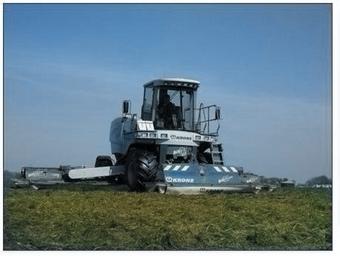}&
        \includegraphics[width=.190\textwidth]{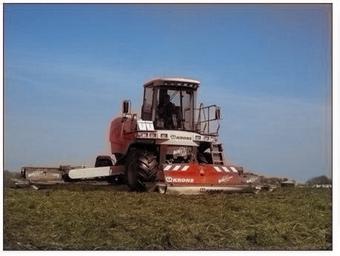}&
        \includegraphics[width=.190\textwidth]{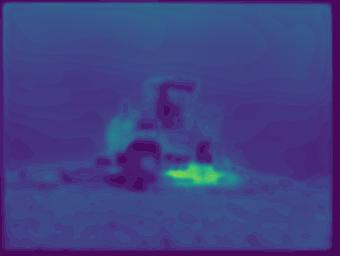}\\
    \end{tabular}
    \caption{
        \small
        \textbf{Sampling multiple colorizations.}
        From left:
            graylevel input;
            three colorizations sampled from our model;
            color uncertainty map according to our model.
    }
    \label{fig:supp-warhol}
\end{figure}

\begin{figure}[!th]
    \RawFloats
    \setlength\fboxsep{0pt}
    \begin{center}
        \begin{minipage}[t]{0.1625\linewidth}
        \begin{center}
            \begin{minipage}[t]{0.975\linewidth}
            \begin{center}
                \includegraphics[width=\textwidth]{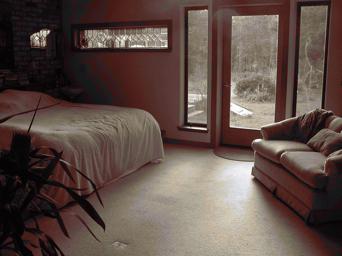}\\
                \includegraphics[width=\textwidth]{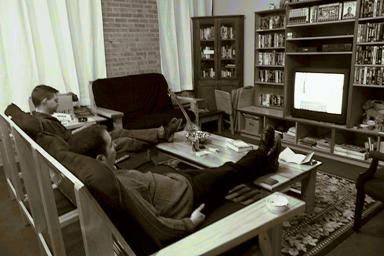}\\
                \includegraphics[width=\textwidth]{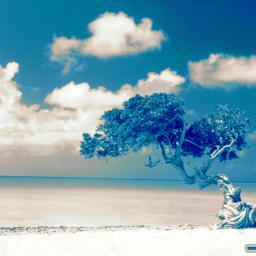}\\
                \includegraphics[width=\textwidth]{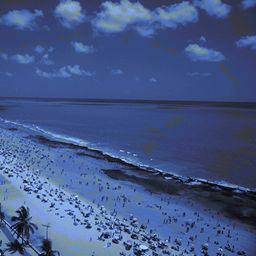}\\
                \includegraphics[width=\textwidth]{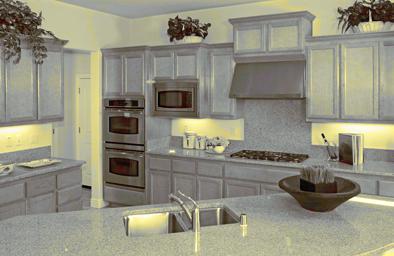}\\
                \includegraphics[width=\textwidth]{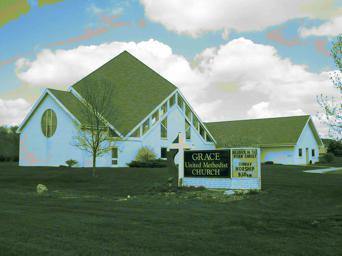}\\
                \scriptsize{\textbf{\textsf{Grayscale only}}}
            \end{center}
            \end{minipage}\\
            \rule{0.95\linewidth}{0.5pt}\\
            \scriptsize{\textbf{\textsf{Welsh~\etal~\cite{welsh2002transferring}}}}
        \end{center}
        \end{minipage}
        \hfill
        \begin{minipage}[t]{0.325\linewidth}
        \begin{center}
            \begin{minipage}[t]{0.4875\linewidth}
            \begin{center}
                \includegraphics[width=\textwidth]{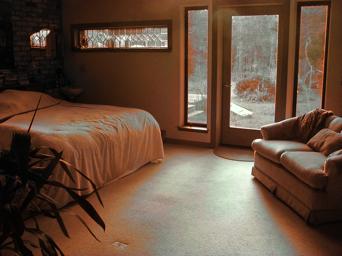}\\
                \includegraphics[width=\textwidth]{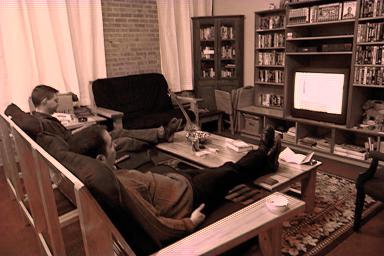}\\
                \includegraphics[width=\textwidth]{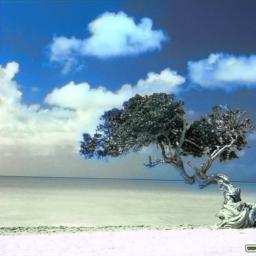}\\
                \includegraphics[width=\textwidth]{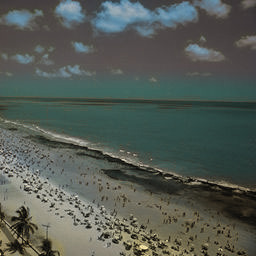}\\
                \includegraphics[width=\textwidth]{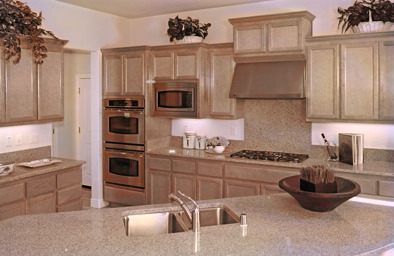}\\
                \includegraphics[width=\textwidth]{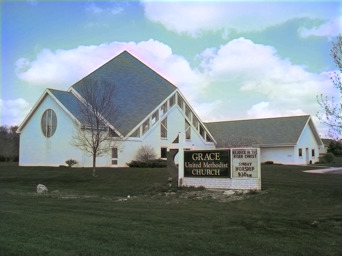}\\
                \scriptsize{\textbf{\textsf{\textcolor{white}{y}GT Scene\textcolor{white}{y}}}}
            \end{center}
            \end{minipage}
            \begin{minipage}[t]{0.4875\linewidth}
            \begin{center}
                \includegraphics[width=\textwidth]{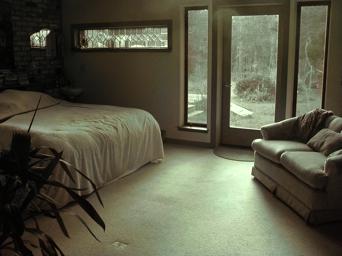}\\
                \includegraphics[width=\textwidth]{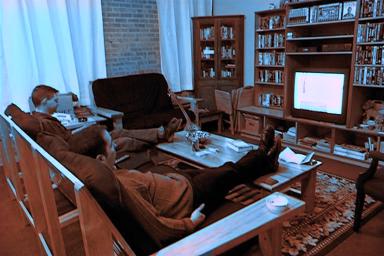}\\
                \includegraphics[width=\textwidth]{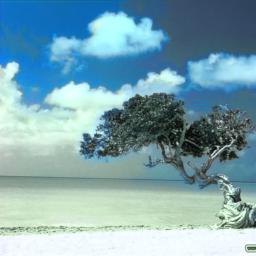}\\
                \includegraphics[width=\textwidth]{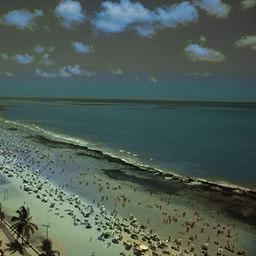}\\
                \includegraphics[width=\textwidth]{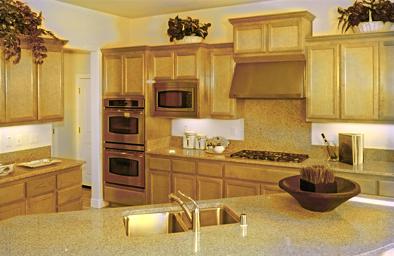}\\
                \includegraphics[width=\textwidth]{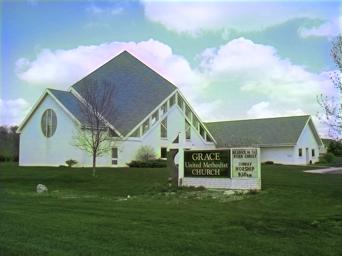}\\
                \scriptsize{\textbf{\textsf{GT Scene \& Hist}}}
            \end{center}
            \end{minipage}\\
            \rule{0.975\linewidth}{0.5pt}\\
            \scriptsize{\textbf{\textsf{Deshpande~\etal~\cite{deshpande2015learning}}}}
        \end{center}
        \end{minipage}
        \hfill
        \begin{minipage}[t]{0.325\linewidth}
        \begin{center}
            \begin{minipage}[t]{0.4875\linewidth}
            \begin{center}
                \includegraphics[width=\textwidth]{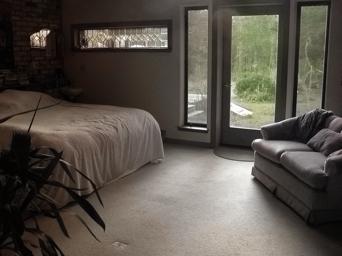}\\
                \includegraphics[width=\textwidth]{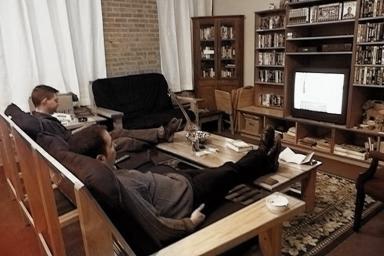}\\
                \includegraphics[width=\textwidth]{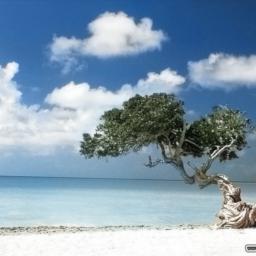}\\
                \includegraphics[width=\textwidth]{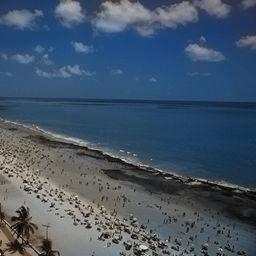}\\
                \includegraphics[width=\textwidth]{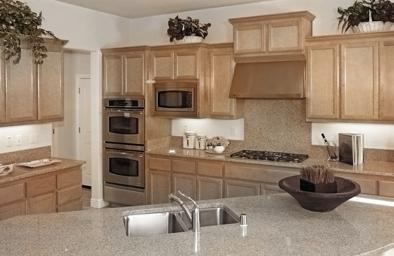}\\
                \includegraphics[width=\textwidth]{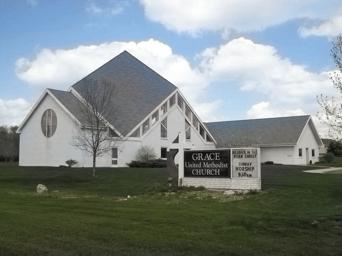}\\
                \scriptsize{\textbf{\textsf{Grayscale only}}}
            \end{center}
            \end{minipage}
            \begin{minipage}[t]{0.4875\linewidth}
            \begin{center}
                \includegraphics[width=\textwidth]{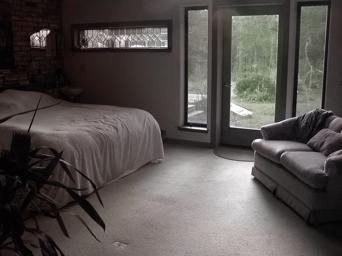}\\
                \includegraphics[width=\textwidth]{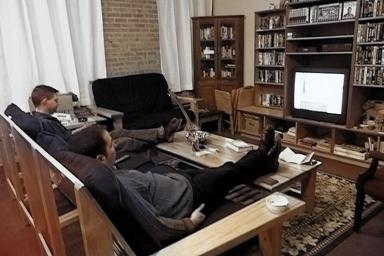}\\
                \includegraphics[width=\textwidth]{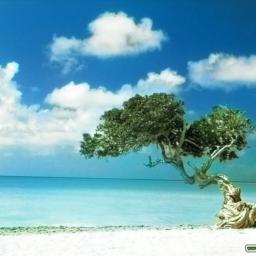}\\
                \includegraphics[width=\textwidth]{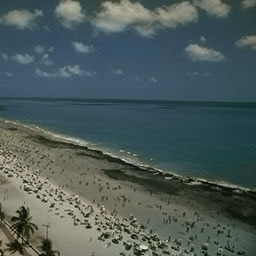}\\
                \includegraphics[width=\textwidth]{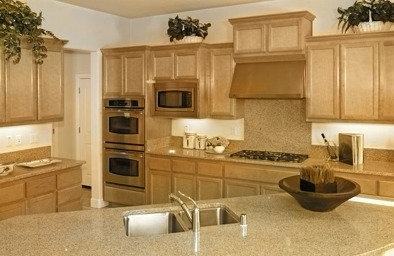}\\
                \includegraphics[width=\textwidth]{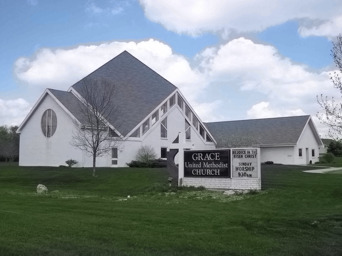}\\
                \scriptsize{\textbf{\textsf{GT Histogram}}}
            \end{center}
            \end{minipage}\\
            \rule{0.975\linewidth}{0.5pt}\\
            \scriptsize{\textbf{\textsf{Our Method}}}
        \end{center}
        \end{minipage}
        \hfill
        \begin{minipage}[t]{0.1625\linewidth}
        \begin{center}
            \begin{minipage}[t]{0.975\linewidth}
            \begin{center}
                \includegraphics[width=\textwidth]{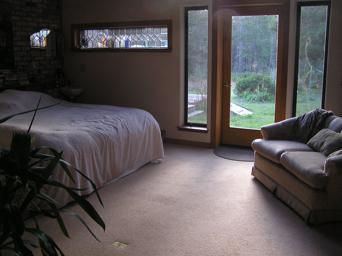}\\
                \includegraphics[width=\textwidth]{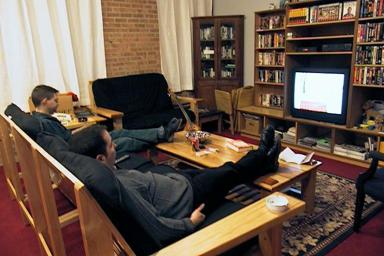}\\
                \includegraphics[width=\textwidth]{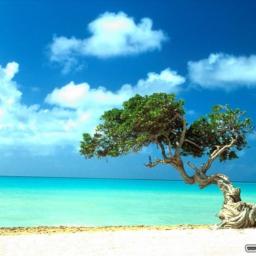}\\
                \includegraphics[width=\textwidth]{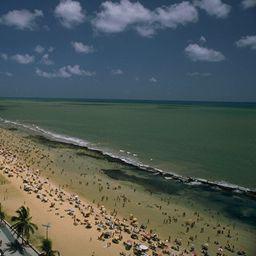}\\
                \includegraphics[width=\textwidth]{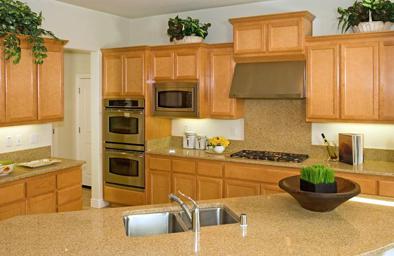}\\
                \includegraphics[width=\textwidth]{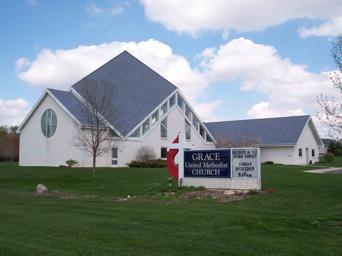}\\
                \scriptsize{\textbf{\textsf{Ground-truth}}}
            \end{center}
            \end{minipage}
        \end{center}
        \end{minipage}
    \end{center}
    \caption{\small 
        \textbf{SUN-6.} Additional qualitative comparisons.
    }
    \label{fig:supp-sun6-examples}
\end{figure}

\begin{figure}[b!]
    \begin{center}
    \begin{minipage}[b]{0.270\linewidth}
        \begin{center}
            \includegraphics[height=2.3cm]{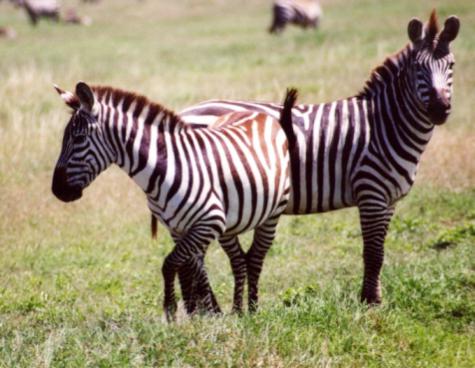}
            \includegraphics[height=1.8cm]{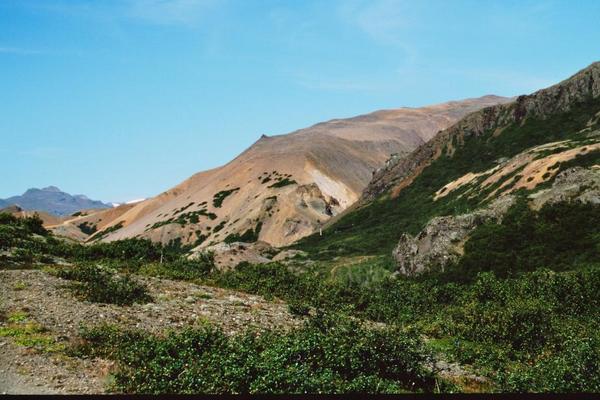}
            \scriptsize{\textbf{\textsf{\ypad{}Reference Image\ypad\\ \ypad}}}
        \end{center}
    \end{minipage}
    \begin{minipage}[b]{0.235\linewidth}
        \begin{center}
            \includegraphics[height=2.3cm]{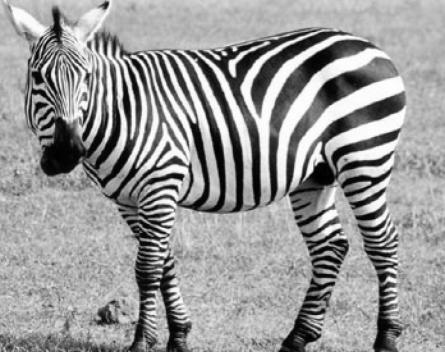}
            \includegraphics[height=1.8cm]{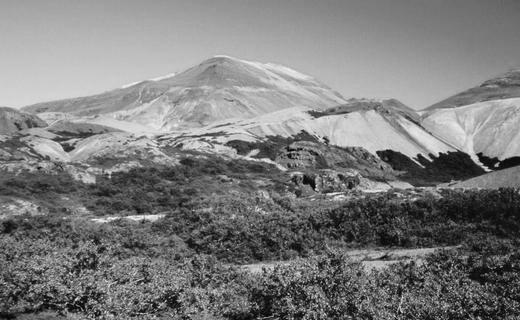}
            \scriptsize{\textbf{\textsf{\ypad{}Input\ypad\\ \ypad}}}
        \end{center}
    \end{minipage}
    \begin{minipage}[b]{0.235\linewidth}
        \begin{center}
            \includegraphics[height=2.3cm]{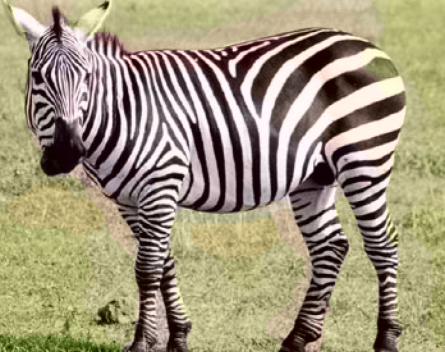}
            \includegraphics[height=1.8cm]{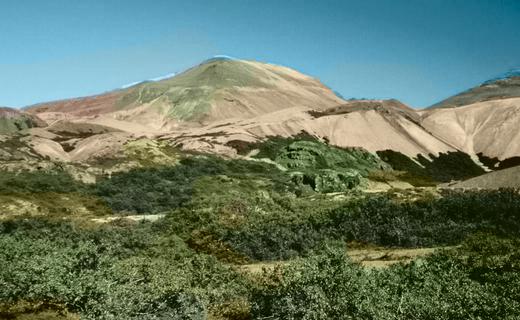}
            \scriptsize{\textbf{\textsf{\ypad{}Charpiat~\etal~\cite{charpiat2010machine}\ypad{}\\ \ypad}}}
        \end{center}
    \end{minipage}
    \begin{minipage}[b]{0.235\linewidth}
        \begin{center}
            \includegraphics[height=2.3cm]{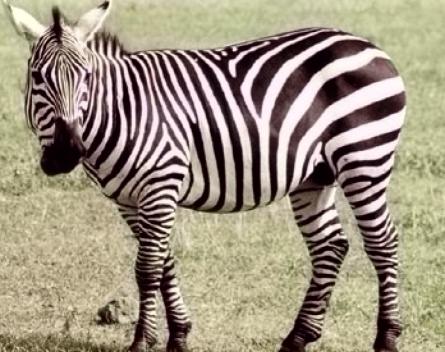}
            \includegraphics[height=1.8cm]{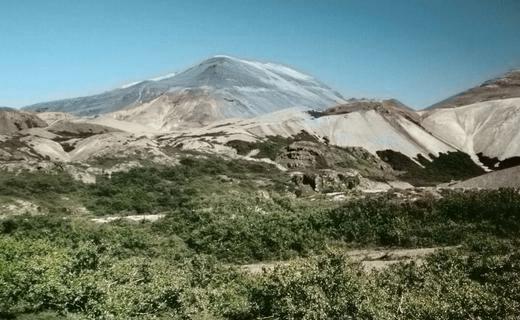}
            \scriptsize{\textbf{\textsf{\ypad{}Our Method\ypad{}\\ (Energy Minimization)}}}
        \end{center}
    \end{minipage}
\end{center}
\caption{
\textbf{Transfer.}
Comparison with Charpiat~\etal~\cite{charpiat2010machine} with reference image. Their method works fairly well when the reference image closely matches (compare with Figure~\ref{fig:charpiat-portraits}). However, they still present sharp unnatural color edges.
We apply our histogram transfer method (Energy Minimization) using the reference image.
}
\label{fig:charpiat-reference}
\end{figure}

\begin{figure}[h!]
    \begin{center}
    \begin{minipage}[b]{0.180\linewidth}
        \begin{center}
            \includegraphics[width=\textwidth]{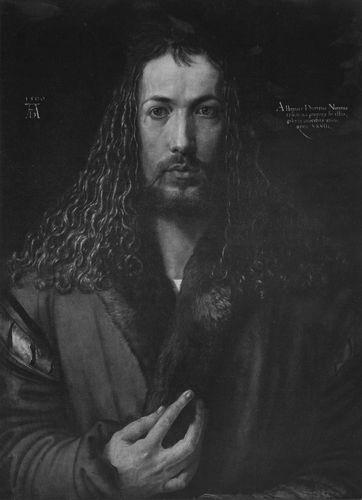}
            \includegraphics[width=\textwidth]{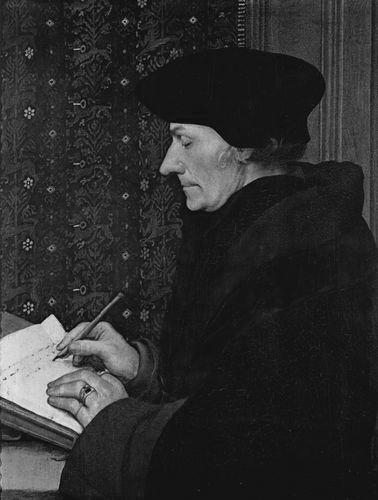}
            \includegraphics[width=\textwidth]{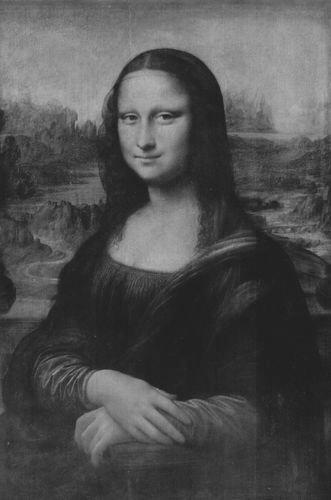}
            \includegraphics[width=\textwidth]{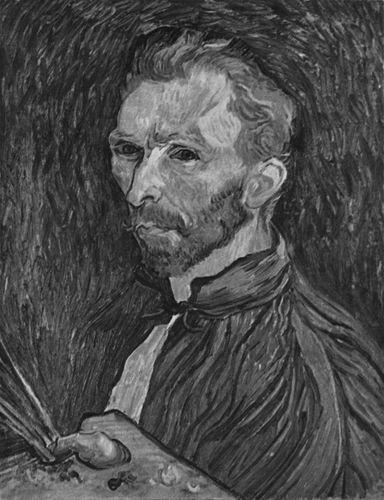}
            \includegraphics[width=\textwidth]{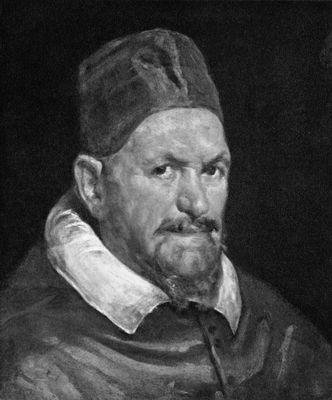}
            \scriptsize{\textbf{\textsf{\phantom{y[}Input}\phantom{k[}}}
        \end{center}
    \end{minipage}
    \begin{minipage}[b]{0.180\linewidth}
        \begin{center}
            \includegraphics[width=\textwidth]{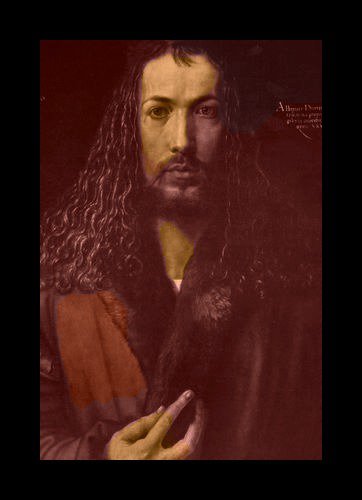}
            \includegraphics[width=\textwidth]{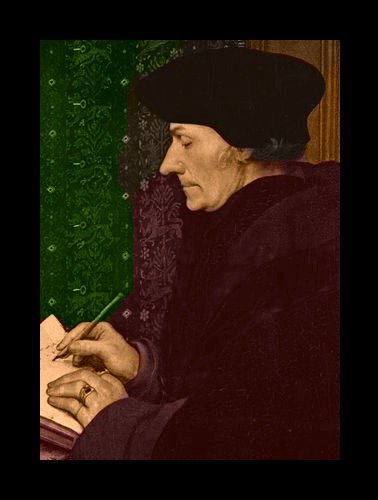}
            \includegraphics[width=\textwidth]{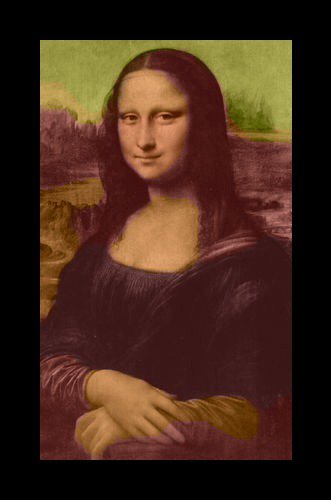}
            \includegraphics[width=\textwidth]{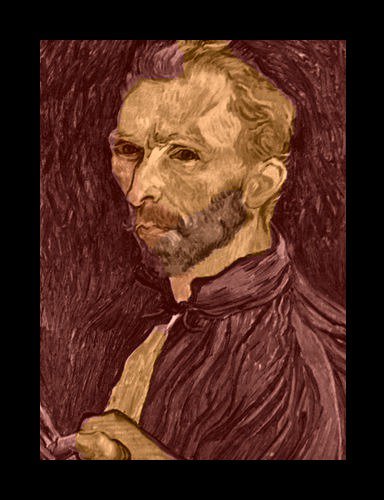}
            \includegraphics[width=\textwidth]{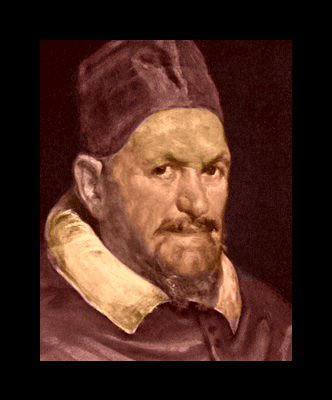}
            \scriptsize{\textbf{\textsf{Charpiat~\etal~\cite{charpiat2010machine}}}}
        \end{center}
    \end{minipage}
    \begin{minipage}[b]{0.180\linewidth}
        \begin{center}
            \includegraphics[width=\textwidth]{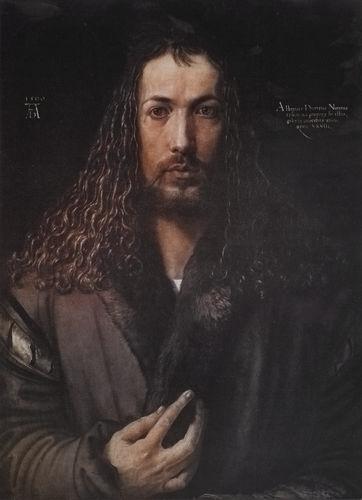}
            \includegraphics[width=\textwidth]{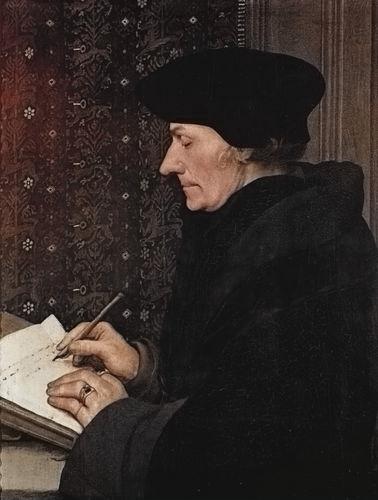}
            \includegraphics[width=\textwidth]{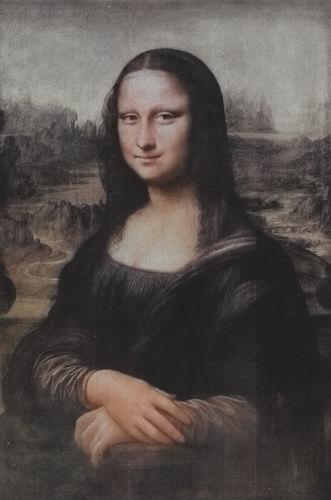}
            \includegraphics[width=\textwidth]{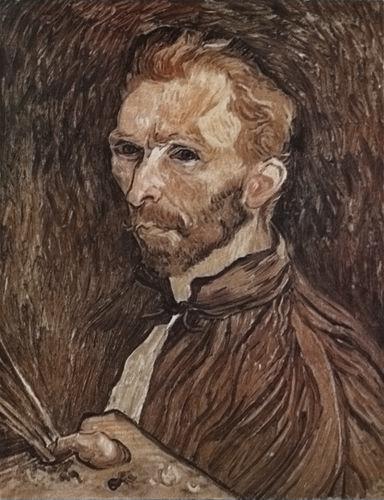}
            \includegraphics[width=\textwidth]{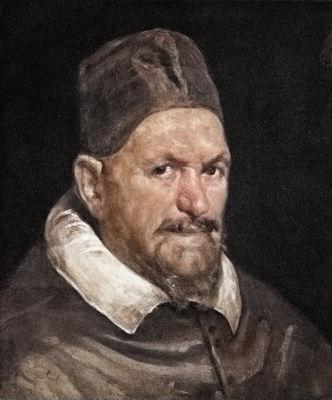}
            \scriptsize{\textbf{\textsf{\phantom{y[}Our Method\phantom{y[}}}}
        \end{center}
    \end{minipage}
    \begin{minipage}[b]{0.180\linewidth}
        \begin{center}
            \includegraphics[width=\textwidth]{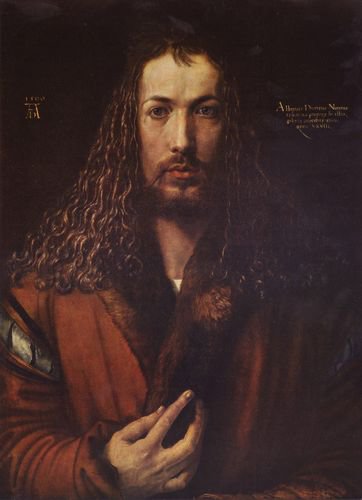}
            \includegraphics[width=\textwidth]{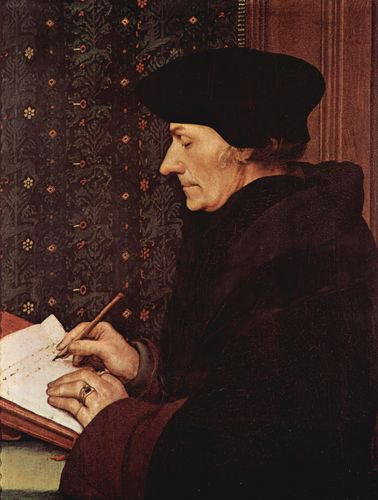}
            \includegraphics[width=\textwidth]{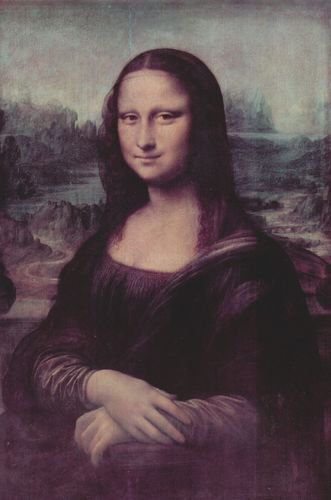}
            \includegraphics[width=\textwidth]{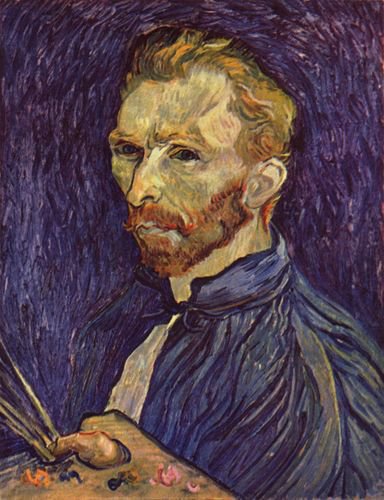}
            \includegraphics[width=\textwidth]{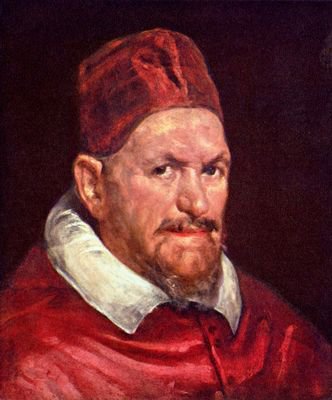}
            \scriptsize{\textbf{\textsf{\phantom{y[}Ground-truth\phantom{y[}}}}
        \end{center}
\end{minipage}
\end{center}
\caption{
\textbf{Portraits.}
Comparison with Charpiat~\etal~\cite{charpiat2010machine}, a transfer-based method using 53 reference portrait paintings. Note that their method works significantly worse when the reference images are not hand-picked for each grayscale input (compare with Figure~\ref{fig:charpiat-reference}). Our model was not trained specifically for this task and we used no reference images.
}
\label{fig:charpiat-portraits}
\end{figure}

\begin{figure}[h!]
    \begin{center}
    \begin{minipage}[b]{0.239\linewidth}
        \vspace{0pt}
        \begin{center}
           \includegraphics[width=\textwidth]{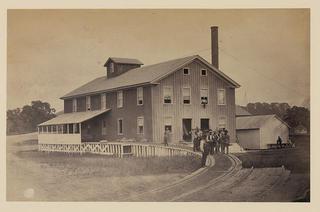}
           \vspace{0.022cm}
           \includegraphics[width=\textwidth]{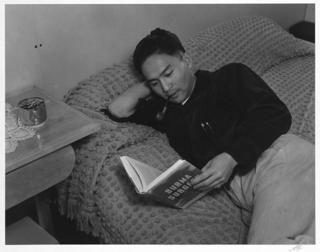}
           \vspace{0.022cm}
           \includegraphics[width=\textwidth]{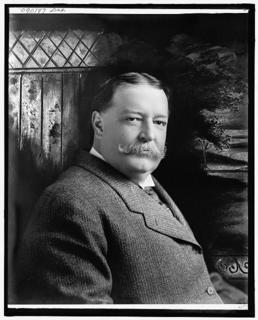}
           \vspace{0.022cm}
           \includegraphics[width=\textwidth]{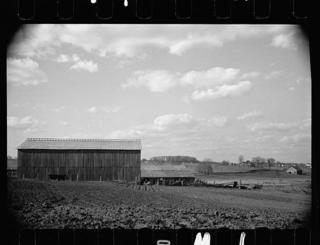}
           \vspace{0.022cm}
           \includegraphics[width=\textwidth]{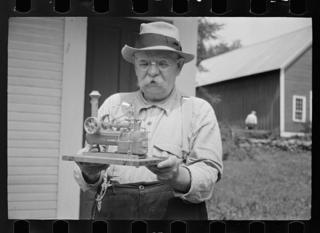}
           \vspace{0.022cm}
           \includegraphics[width=\textwidth]{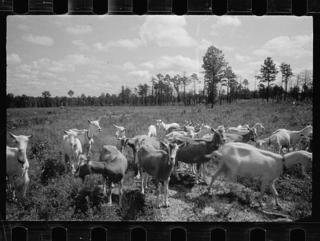}
            \vspace{0.02\linewidth}
            \scriptsize{\textbf{\textsf{\phantom{y[}Input\phantom{y[}}}}
        \end{center}
    \end{minipage}
    \begin{minipage}[b]{0.239\linewidth}
        \vspace{0pt}
        \begin{center}
           \includegraphics[width=\textwidth]{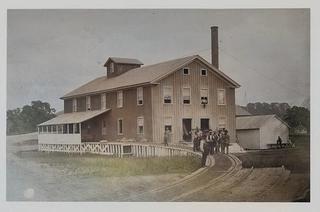}
           \vspace{0.022cm}
           \includegraphics[width=\textwidth]{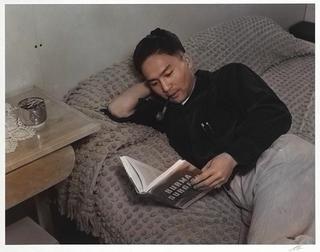}
           \vspace{0.022cm}
           \includegraphics[width=\textwidth]{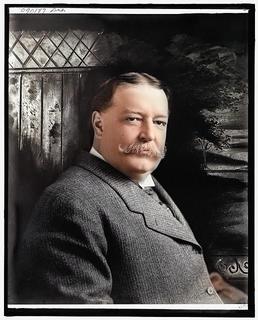}
           \vspace{0.022cm}
           \includegraphics[width=\textwidth]{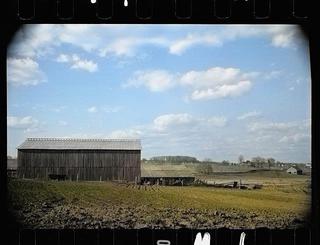}
           \vspace{0.022cm}
           \includegraphics[width=\textwidth]{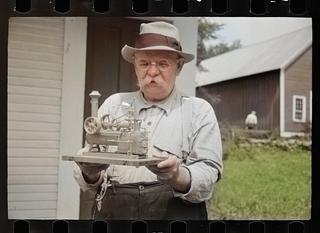}
           \vspace{0.022cm}
           \includegraphics[width=\textwidth]{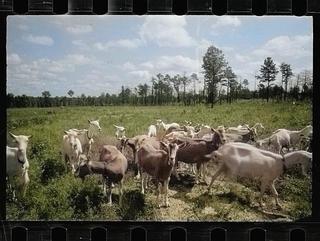}
            \vspace{0.02\linewidth}
            \scriptsize{\textbf{\textsf{\phantom{y[}Our Method\phantom{y[}}}}
        \end{center}
    \end{minipage}
    ~
    \begin{minipage}[b]{0.239\linewidth}
        \vspace{0pt}
        \begin{center}
           \includegraphics[width=\textwidth]{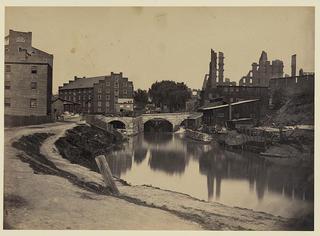}
           \includegraphics[width=\textwidth]{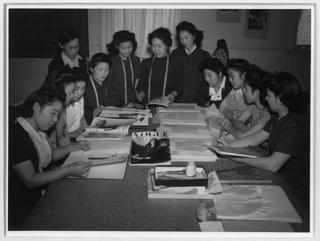}
           \includegraphics[width=\textwidth]{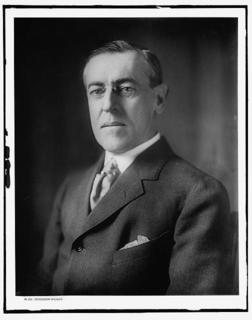}
           \includegraphics[width=\textwidth]{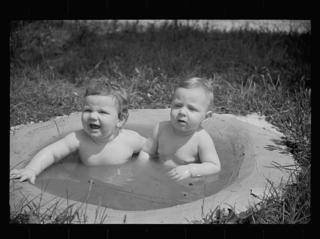}
           \includegraphics[width=\textwidth]{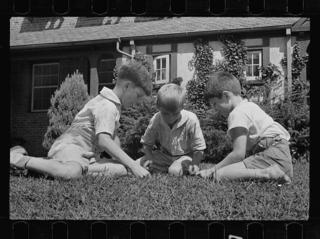}
           \includegraphics[width=\textwidth]{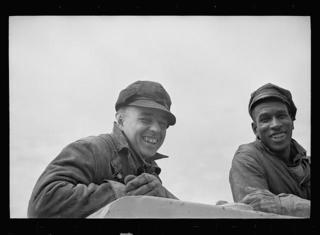}
            \vspace{0.02\linewidth}
            \scriptsize{\textbf{\textsf{\phantom{y[}Input\phantom{y[}}}}
        \end{center}
    \end{minipage}
    \begin{minipage}[b]{0.239\linewidth}
        \vspace{0pt}
        \begin{center}
           \includegraphics[width=\textwidth]{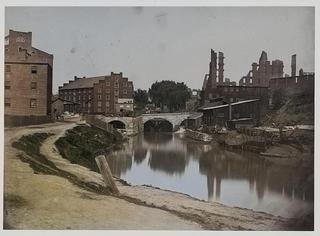}
           \includegraphics[width=\textwidth]{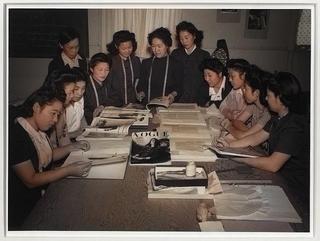}
           \includegraphics[width=\textwidth]{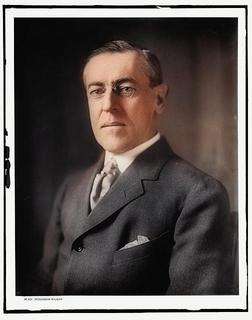}
           \includegraphics[width=\textwidth]{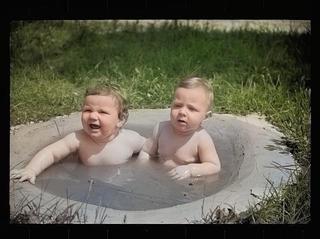}
           \includegraphics[width=\textwidth]{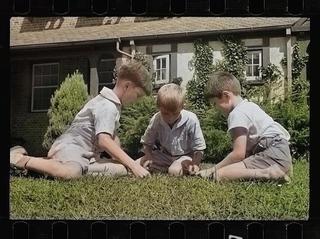}
           \includegraphics[width=\textwidth]{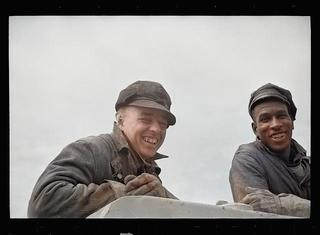}
           \vspace{0.02\linewidth}
           \scriptsize{\textbf{\textsf{\phantom{y[}Our Method\phantom{y[}}}}
        \end{center}
\end{minipage}
\end{center}
    \caption{
        \textbf{B\&W photographs.} Old photographs that were automatically colorized. (Source: Library of Congress, \texttt{www.loc.gov})
    }
    \label{fig:supp-legacy}
\end{figure}

We provide additional samples for global biasing (Figure~\ref{fig:supp-warhol}) and SUN-6 (Figure~\ref{fig:supp-sun6-examples}). Comparisons with Charpiat~\etal~\cite{charpiat2010machine} appear in Figures~\ref{fig:charpiat-reference} and~\ref{fig:charpiat-portraits}. Examples of how our algorithm can bring old photographs to life in Figure~\ref{fig:supp-legacy}. More examples on ImageNet (ctest10k) in Figures~\ref{fig:ctest10k-supp-examples1} to~\ref{fig:ctest10k-supp-examples4}
%,fig:ctest10k-supp-examples2,fig:ctest10k-supp-examples3,fig:ctest10k-supp-examples4} 
and Figure~\ref{fig:ctest10k-supp-failures1} (failure cases). Examples of histogram predictions in Figures~\ref{fig:histogram-supp-field} and~\ref{fig:histogram-supp-parrot}.

\begin{figure}[!th]
    \begin{center}
    \begin{minipage}[b]{0.158\linewidth}
        \vspace{0pt}
        \begin{center}
            \includegraphics[width=\textwidth]{\gs{41970}}\\
            \includegraphics[width=\textwidth]{\gs{41918}}\\
            \includegraphics[width=\textwidth]{\gs{41497}}\\
            \includegraphics[width=\textwidth]{\gs{41928}}\\
            \includegraphics[width=\textwidth]{\gs{41937}}\\
            \includegraphics[width=\textwidth]{\gs{41939}}\\
            \includegraphics[width=\textwidth]{\gs{41958}}\\
            \includegraphics[width=\textwidth]{\gs{42718}}\\
            \includegraphics[width=\textwidth]{\gs{42729}}\\
            \includegraphics[width=\textwidth]{\gs{42085}}\\
            \vspace{0.02\linewidth}
            \scriptsize{\textbf{\textsf{Input}}}
        \end{center}
    \end{minipage}
    \begin{minipage}[b]{0.158\linewidth}
        \vspace{0pt}
        \begin{center}
            \includegraphics[width=\textwidth]{\mm{41970}}\\
            \includegraphics[width=\textwidth]{\mm{41918}}\\
            \includegraphics[width=\textwidth]{\mm{41497}}\\
            \includegraphics[width=\textwidth]{\mm{41928}}\\
            \includegraphics[width=\textwidth]{\mm{41937}}\\
            \includegraphics[width=\textwidth]{\mm{41939}}\\
            \includegraphics[width=\textwidth]{\mm{41958}}\\
            \includegraphics[width=\textwidth]{\mm{42718}}\\
            \includegraphics[width=\textwidth]{\mm{42729}}\\
            \includegraphics[width=\textwidth]{\mm{42085}}\\
            \vspace{0.02\linewidth}
            \scriptsize{\textbf{\textsf{Our Method}}}
        \end{center}
    \end{minipage}
    \begin{minipage}[b]{0.158\linewidth}
        \vspace{0pt}
        \begin{center}
            \includegraphics[width=\textwidth]{\gt{41970}}\\
            \includegraphics[width=\textwidth]{\gt{41918}}\\
            \includegraphics[width=\textwidth]{\gt{41497}}\\
            \includegraphics[width=\textwidth]{\gt{41928}}\\
            \includegraphics[width=\textwidth]{\gt{41937}}\\
            \includegraphics[width=\textwidth]{\gt{41939}}\\
            \includegraphics[width=\textwidth]{\gt{41958}}\\
            \includegraphics[width=\textwidth]{\gt{42718}}\\
            \includegraphics[width=\textwidth]{\gt{42729}}\\
            \includegraphics[width=\textwidth]{\gt{42085}}\\
            \vspace{0.02\linewidth}
            \scriptsize{\textbf{\textsf{Ground-truth}}}
        \end{center}
    \end{minipage}
    \hfill
    \begin{minipage}[b]{0.158\linewidth}
        \vspace{0pt}
        \begin{center}
            \includegraphics[width=\textwidth]{\gs{42700}}\\
            \vspace{0.033cm}
            \includegraphics[width=\textwidth]{\gs{40375}}\\
            \vspace{0.033cm}
            \includegraphics[width=\textwidth]{\gs{41979}}\\
            \vspace{0.033cm}
            \includegraphics[width=\textwidth]{\gs{41995}}\\
            \vspace{0.033cm}
            \includegraphics[width=\textwidth]{\gs{42010}}\\
            \vspace{0.033cm}
            \includegraphics[width=\textwidth]{\gs{42022}}\\
            \vspace{0.033cm}
            \includegraphics[width=\textwidth]{\gs{42054}}\\
            \vspace{0.033cm}
            \includegraphics[width=\textwidth]{\gs{42093}}\\
            \vspace{0.033cm}
            \includegraphics[width=\textwidth]{\gs{42133}}\\
            \vspace{0.02\linewidth}
            \scriptsize{\textbf{\textsf{Input}}}
        \end{center}
    \end{minipage}
    \begin{minipage}[b]{0.158\linewidth}
        \vspace{0pt}
        \begin{center}
            \includegraphics[width=\textwidth]{\mm{42700}}\\
            \vspace{0.033cm}
            \includegraphics[width=\textwidth]{\mm{40375}}\\
            \vspace{0.033cm}
            \includegraphics[width=\textwidth]{\mm{41979}}\\
            \vspace{0.033cm}
            \includegraphics[width=\textwidth]{\mm{41995}}\\
            \vspace{0.033cm}
            \includegraphics[width=\textwidth]{\mm{42010}}\\
            \vspace{0.033cm}
            \includegraphics[width=\textwidth]{\mm{42022}}\\
            \vspace{0.033cm}
            \includegraphics[width=\textwidth]{\mm{42054}}\\
            \vspace{0.033cm}
            \includegraphics[width=\textwidth]{\mm{42093}}\\
            \vspace{0.033cm}
            \includegraphics[width=\textwidth]{\mm{42133}}\\
            \vspace{0.02\linewidth}
            \scriptsize{\textbf{\textsf{Our Method}}}
        \end{center}
    \end{minipage}
    \begin{minipage}[b]{0.158\linewidth}
        \vspace{0pt}
        \begin{center}
            \includegraphics[width=\textwidth]{\gt{42700}}\\
            \vspace{0.033cm}
            \includegraphics[width=\textwidth]{\gt{40375}}\\
            \vspace{0.033cm}
            \includegraphics[width=\textwidth]{\gt{41979}}\\
            \vspace{0.033cm}
            \includegraphics[width=\textwidth]{\gt{41995}}\\
            \vspace{0.033cm}
            \includegraphics[width=\textwidth]{\gt{42010}}\\
            \vspace{0.033cm}
            \includegraphics[width=\textwidth]{\gt{42022}}\\
            \vspace{0.033cm}
            \includegraphics[width=\textwidth]{\gt{42054}}\\
            \vspace{0.033cm}
            \includegraphics[width=\textwidth]{\gt{42093}}\\
            \vspace{0.033cm}
            \includegraphics[width=\textwidth]{\gt{42133}}\\
            \vspace{0.02\linewidth}
            \scriptsize{\textbf{\textsf{Ground-truth}}}
        \end{center}
    \end{minipage}
    \end{center}
    \caption{\small 
        \textbf{Fully automatic colorization results on ImageNet/ctest10k.}
    }
    \label{fig:ctest10k-supp-examples1}
\end{figure}

\begin{figure}[!th]
    \begin{center}
    \begin{minipage}[b]{0.158\linewidth}
        \vspace{0pt}
        \begin{center}
            \includegraphics[width=\textwidth]{\gs{42488}}\\
            \includegraphics[width=\textwidth]{\gs{42444}}\\
            \includegraphics[width=\textwidth]{\gs{42151}}\\
            \includegraphics[width=\textwidth]{\gs{42697}}\\
            \includegraphics[width=\textwidth]{\gs{42698}}\\
            \includegraphics[width=\textwidth]{\gs{42699}}\\
            \includegraphics[width=\textwidth]{\gs{43987}}\\
            \includegraphics[width=\textwidth]{\gs{44100}}\\
            \vspace{0.02\linewidth}
            \scriptsize{\textbf{\textsf{Input}}}
        \end{center}
    \end{minipage}
    \begin{minipage}[b]{0.158\linewidth}
        \vspace{0pt}
        \begin{center}
            \includegraphics[width=\textwidth]{\mm{42488}}\\
            \includegraphics[width=\textwidth]{\mm{42444}}\\
            \includegraphics[width=\textwidth]{\mm{42151}}\\
            \includegraphics[width=\textwidth]{\mm{42697}}\\
            \includegraphics[width=\textwidth]{\mm{42698}}\\
            \includegraphics[width=\textwidth]{\mm{42699}}\\
            \includegraphics[width=\textwidth]{\mm{43987}}\\
            \includegraphics[width=\textwidth]{\mm{44100}}\\
            \vspace{0.02\linewidth}
            \scriptsize{\textbf{\textsf{Our Method}}}
        \end{center}
    \end{minipage}
    \begin{minipage}[b]{0.158\linewidth}
        \vspace{0pt}
        \begin{center}
            \includegraphics[width=\textwidth]{\gt{42488}}\\
            \includegraphics[width=\textwidth]{\gt{42444}}\\
            \includegraphics[width=\textwidth]{\gt{42151}}\\
            \includegraphics[width=\textwidth]{\gt{42697}}\\
            \includegraphics[width=\textwidth]{\gt{42698}}\\
            \includegraphics[width=\textwidth]{\gt{42699}}\\
            \includegraphics[width=\textwidth]{\gt{43987}}\\
            \includegraphics[width=\textwidth]{\gt{44100}}\\
            \vspace{0.02\linewidth}
            \scriptsize{\textbf{\textsf{Ground-truth}}}
        \end{center}
    \end{minipage}
    \hfill
    \begin{minipage}[b]{0.158\linewidth}
        \vspace{0pt}
        \begin{center}
            \includegraphics[width=\textwidth]{\gs{42451}}\\
            \includegraphics[width=\textwidth]{\gs{42481}}\\
            \includegraphics[width=\textwidth]{\gs{42554}}\\
            \includegraphics[width=\textwidth]{\gs{42153}}\\
            \includegraphics[width=\textwidth]{\gs{42165}}\\
            \includegraphics[width=\textwidth]{\gs{42194}}\\
            \includegraphics[width=\textwidth]{\gs{42207}}\\
            \includegraphics[width=\textwidth]{\gs{42300}}\\
            \includegraphics[width=\textwidth]{\gs{42358}}\\
            \includegraphics[width=\textwidth]{\gs{43959}}\\
            \includegraphics[width=\textwidth]{\gs{43985}}\\
            \vspace{0.02\linewidth}
            \scriptsize{\textbf{\textsf{Input}}}
        \end{center}
    \end{minipage}
    \begin{minipage}[b]{0.158\linewidth}
        \vspace{0pt}
        \begin{center}
            \includegraphics[width=\textwidth]{\mm{42451}}\\
            \includegraphics[width=\textwidth]{\mm{42481}}\\
            \includegraphics[width=\textwidth]{\mm{42554}}\\
            \includegraphics[width=\textwidth]{\mm{42153}}\\
            \includegraphics[width=\textwidth]{\mm{42165}}\\
            \includegraphics[width=\textwidth]{\mm{42194}}\\
            \includegraphics[width=\textwidth]{\mm{42207}}\\
            \includegraphics[width=\textwidth]{\mm{42300}}\\
            \includegraphics[width=\textwidth]{\mm{42358}}\\
            \includegraphics[width=\textwidth]{\mm{43959}}\\
            \includegraphics[width=\textwidth]{\mm{43985}}\\
            \vspace{0.02\linewidth}
            \scriptsize{\textbf{\textsf{Our Method}}}
        \end{center}
    \end{minipage}
    \begin{minipage}[b]{0.158\linewidth}
        \vspace{0pt}
        \begin{center}
            \includegraphics[width=\textwidth]{\gt{42451}}\\
            \includegraphics[width=\textwidth]{\gt{42481}}\\
            \includegraphics[width=\textwidth]{\gt{42554}}\\
            \includegraphics[width=\textwidth]{\gt{42153}}\\
            \includegraphics[width=\textwidth]{\gt{42165}}\\
            \includegraphics[width=\textwidth]{\gt{42194}}\\
            \includegraphics[width=\textwidth]{\gt{42207}}\\
            \includegraphics[width=\textwidth]{\gt{42300}}\\
            \includegraphics[width=\textwidth]{\gt{42358}}\\
            \includegraphics[width=\textwidth]{\gt{43959}}\\
            \includegraphics[width=\textwidth]{\gt{43985}}\\
            \vspace{0.02\linewidth}
            \scriptsize{\textbf{\textsf{Ground-truth}}}
        \end{center}
    \end{minipage}
    \end{center}
    \caption{\small 
        \textbf{Fully automatic colorization results on ImageNet/ctest10k.}
    }
    \label{fig:ctest10k-supp-examples2}
\end{figure}

\begin{figure}[!th]
    \begin{center}
    \begin{minipage}[b]{0.19\linewidth}
        \vspace{0pt}
        \begin{center}
            \includegraphics[width=\textwidth]{\mm{41064}}\\
            \vspace{0.027cm}
            \includegraphics[width=\textwidth]{\mm{37073}}\\
            \vspace{0.027cm}
            \includegraphics[width=\textwidth]{\mm{37123}}\\
            \vspace{0.027cm}
            \includegraphics[width=\textwidth]{\mm{37215}}\\
            \vspace{0.027cm}
            \includegraphics[width=\textwidth]{\mm{37538}}\\
            \vspace{0.027cm}
            \includegraphics[width=\textwidth]{\mm{37379}}\\
            \vspace{0.027cm}
            \includegraphics[width=\textwidth]{\mm{37247}}\\
            \vspace{0.027cm}
            \includegraphics[width=\textwidth]{\mm{42817}}\\
            \vspace{0.027cm}
            \includegraphics[width=\textwidth]{\mm{37129}}\\
        \end{center}
    \end{minipage}
    \hfill
    \begin{minipage}[b]{0.19\linewidth}
        \vspace{0.10\linewidth}
        \begin{center}
            \includegraphics[width=\textwidth]{\mm{38651}}\\
            \vspace{0.002cm}
            \includegraphics[width=\textwidth]{\mm{38734}}\\
            \vspace{0.002cm}
            \includegraphics[width=\textwidth]{\mm{37404}}\\
            \vspace{0.002cm}
            \includegraphics[width=\textwidth]{\mm{37468}}\\
            \vspace{0.002cm}
            \includegraphics[width=\textwidth]{\mm{37619}}\\
            \vspace{0.002cm}
            \includegraphics[width=\textwidth]{\mm{42942}}\\
            \vspace{0.002cm}
            \includegraphics[width=\textwidth]{\mm{38739}}\\
            \vspace{0.002cm}
            \includegraphics[width=\textwidth]{\mm{37222}}\\
        \end{center}
    \end{minipage}
    \hfill
    \begin{minipage}[b]{0.19\linewidth}
        \vspace{0pt}
        \begin{center}
            \includegraphics[width=\textwidth]{\mm{37813}}\\
            \vspace{0.005cm}
            \includegraphics[width=\textwidth]{\mm{43649}}\\
            \vspace{0.005cm}
            \includegraphics[width=\textwidth]{\mm{43863}}\\
            \vspace{0.005cm}
            \includegraphics[width=\textwidth]{\mm{37552}}\\
            \vspace{0.005cm}
            \includegraphics[width=\textwidth]{\mm{43529}}\\
            \vspace{0.005cm}
            \includegraphics[width=\textwidth]{\mm{37658}}\\
            \vspace{0.005cm}
            \includegraphics[width=\textwidth]{\mm{37736}}\\
            \vspace{0.005cm}
            \includegraphics[width=\textwidth]{\mm{37749}}\\
            \vspace{0.005cm}
            \includegraphics[width=\textwidth]{\mm{37522}}\\
        \end{center}
    \end{minipage}
    \hfill
    \begin{minipage}[b]{0.19\linewidth}
        \vspace{0pt}
        \begin{center}
            \includegraphics[width=\textwidth]{\mm{43638}}\\
            \includegraphics[width=\textwidth]{\mm{37915}}\\
            \includegraphics[width=\textwidth]{\mm{37980}}\\
            \includegraphics[width=\textwidth]{\mm{38061}}\\
            \includegraphics[width=\textwidth]{\mm{38209}}\\
            \includegraphics[width=\textwidth]{\mm{38273}}\\
            \includegraphics[width=\textwidth]{\mm{39170}}\\
            \includegraphics[width=\textwidth]{\mm{43178}}\\
        \end{center}
    \end{minipage}
    \hfill
    \begin{minipage}[b]{0.19\linewidth}
        \vspace{0pt}
        \begin{center}
            \includegraphics[width=\textwidth]{\mm{43519}}\\
            \vspace{0.03cm}
            \includegraphics[width=\textwidth]{\mm{38756}}\\
            \vspace{0.03cm}
            \includegraphics[width=\textwidth]{\mm{38932}}\\
            \vspace{0.03cm}
            \includegraphics[width=\textwidth]{\mm{38973}}\\
            \vspace{0.03cm}
            \includegraphics[width=\textwidth]{\mm{39089}}\\
            \vspace{0.03cm}
            \includegraphics[width=\textwidth]{\mm{38424}}\\
            \vspace{0.03cm}
            \includegraphics[width=\textwidth]{\mm{39242}}
            \vspace{0.03cm}
            \includegraphics[width=\textwidth]{\mm{43612}}\\
        \end{center}
    \end{minipage}
    \end{center}
    \caption{\small
        \textbf{Fully automatic colorization results on ImageNet/ctest10k.}
    }
    \label{fig:ctest10k-supp-examples3}
\end{figure}

\begin{figure}[!th]
    \begin{center}
    \begin{minipage}[b]{0.19\linewidth}
        \vspace{0pt}
        \begin{center}
            \includegraphics[width=\textwidth]{\mm{39284}}\\
            \vspace{0.02cm}
            \includegraphics[width=\textwidth]{\mm{37780}}\\
            \vspace{0.02cm}
            \includegraphics[width=\textwidth]{\mm{39341}}\\
            \vspace{0.02cm}
            \includegraphics[width=\textwidth]{\mm{39371}}\\
            \vspace{0.02cm}
            \includegraphics[width=\textwidth]{\mm{39950}}\\
            \vspace{0.02cm}
            \includegraphics[width=\textwidth]{\mm{39993}}\\
            \vspace{0.02cm}
            \includegraphics[width=\textwidth]{\mm{40001}}\\
            \vspace{0.02cm}
            \includegraphics[width=\textwidth]{\mm{40021}}\\
            \vspace{0.02cm}
            \includegraphics[width=\textwidth]{\mm{40062}}\\
        \end{center}
    \end{minipage}
    \hfill
    \begin{minipage}[b]{0.19\linewidth}
        \vspace{0.10\linewidth}
        \begin{center}
            \includegraphics[width=\textwidth]{\mm{41852}}\\
            \vspace{0.006cm}
            \includegraphics[width=\textwidth]{\mm{40185}}\\
            \vspace{0.006cm}
            \includegraphics[width=\textwidth]{\mm{40587}}\\
            \vspace{0.006cm}
            \includegraphics[width=\textwidth]{\mm{40542}}\\
            \vspace{0.006cm}
            \includegraphics[width=\textwidth]{\mm{40300}}\\
            \vspace{0.006cm}
            \includegraphics[width=\textwidth]{\mm{40313}}\\
            \vspace{0.006cm}
            \includegraphics[width=\textwidth]{\mm{40363}}\\
            \vspace{0.006cm}
            \includegraphics[width=\textwidth]{\mm{41638}}\\
        \end{center}
    \end{minipage}
    \hfill
    \begin{minipage}[b]{0.19\linewidth}
        \vspace{0pt}
        \begin{center}
            \includegraphics[width=\textwidth]{\mm{40446}}\\
            \vspace{0.002cm}
            \includegraphics[width=\textwidth]{\mm{40191}}\\
            \vspace{0.002cm}
            \includegraphics[width=\textwidth]{\mm{40557}}\\
            \vspace{0.002cm}
            \includegraphics[width=\textwidth]{\mm{40212}}\\
            \vspace{0.002cm}
            \includegraphics[width=\textwidth]{\mm{40622}}\\
            \vspace{0.002cm}
            \includegraphics[width=\textwidth]{\mm{40821}}\\
            \vspace{0.002cm}
            \includegraphics[width=\textwidth]{\mm{40953}}\\
            \vspace{0.002cm}
            \includegraphics[width=\textwidth]{\mm{41863}}\\ 
        \end{center}
    \end{minipage}
    \hfill
    \begin{minipage}[b]{0.19\linewidth}
        \vspace{0pt}
        \begin{center}
            \includegraphics[width=\textwidth]{\mm{41217}}\\
            \vspace{0.013cm}
            \includegraphics[width=\textwidth]{\mm{41272}}\\
            \vspace{0.013cm}
            \includegraphics[width=\textwidth]{\mm{41425}}\\
            \vspace{0.013cm}
            \includegraphics[width=\textwidth]{\mm{41497}}\\
            \vspace{0.013cm}
            \includegraphics[width=\textwidth]{\mm{41544}}\\
            \vspace{0.013cm}
            \includegraphics[width=\textwidth]{\mm{41588}}\\
            \vspace{0.013cm}
            \includegraphics[width=\textwidth]{\mm{41599}}\\
            \vspace{0.013cm}
            \includegraphics[width=\textwidth]{\mm{41611}}\\
            \vspace{0.013cm}
            \includegraphics[width=\textwidth]{\mm{40375}}\\
        \end{center}
    \end{minipage}
    \hfill
    \begin{minipage}[b]{0.19\linewidth}
        \vspace{0pt}
        \begin{center}
            \includegraphics[width=\textwidth]{\mm{40080}}\\
            \vspace{0.015cm}
            \includegraphics[width=\textwidth]{\mm{41672}}\\
            \vspace{0.015cm}
            \includegraphics[width=\textwidth]{\mm{41854}}\\
            \vspace{0.015cm}
            \includegraphics[width=\textwidth]{\mm{40087}}\\
            \vspace{0.015cm}
            \includegraphics[width=\textwidth]{\mm{40161}}\\
            \vspace{0.015cm}
            \includegraphics[width=\textwidth]{\mm{41010}}\\
            \vspace{0.015cm}
            \includegraphics[width=\textwidth]{\mm{40408}}\\
            \vspace{0.015cm}
            \includegraphics[width=\textwidth]{\mm{40410}}\\
        \end{center}
    \end{minipage}
    \end{center}
    \caption{\small
        \textbf{Fully automatic colorization results on ImageNet/ctest10k.}
    }
    \label{fig:ctest10k-supp-examples4}
\end{figure}

\begin{figure}[!th]
    \begin{center}
    \begin{minipage}[b]{0.193\linewidth}
        \vspace{0pt}
        \begin{center}
            % too gray
            \includegraphics[width=\textwidth]{\mm{38007}}\\
            \vspace{0.033cm}
            \includegraphics[width=\textwidth]{\mm{38915}}\\
            \vspace{0.033cm}
            \includegraphics[width=\textwidth]{\mm{39897}}\\
            \vspace{0.033cm}
            \includegraphics[width=\textwidth]{\mm{40016}}\\
            \vspace{0.033cm}
            \includegraphics[width=\textwidth]{\mm{40075}}\\
            \vspace{0.033cm}
            \includegraphics[width=\textwidth]{\mm{40296}}\\
            \vspace{0.033cm}
            \includegraphics[width=\textwidth]{\mm{41662}}\\
            \vspace{0.033cm}
            \includegraphics[width=\textwidth]{\mm{41413}}\\
            \vspace{0.02\linewidth}
            \scriptsize{\textbf{\textsf{Too Desaturated}}}

        \end{center}
    \end{minipage}
    \hfill
    \begin{minipage}[b]{0.193\linewidth}
        \vspace{0pt}
        \begin{center}
            %inconsistent chroma
            \includegraphics[width=\textwidth]{\mm{43187}}\\
            \vspace{0.00cm}
            \includegraphics[width=\textwidth]{\mm{39902}}\\
            \vspace{0.00cm}
            \includegraphics[width=\textwidth]{\mm{38138}}\\
            \vspace{0.00cm}
            \includegraphics[width=\textwidth]{\mm{41631}}\\
            \vspace{0.00cm}
            \includegraphics[width=\textwidth]{\mm{42214}}\\
            \vspace{0.00cm}
            \includegraphics[width=\textwidth]{\mm{37131}}\\
            \vspace{0.00cm}
            \includegraphics[width=\textwidth]{\mm{42949}}\\
            \vspace{0.00cm}
            \includegraphics[width=\textwidth]{\mm{42844}}\\
            \vspace{0.00cm}
            \includegraphics[width=\textwidth]{\mm{43723}}\\
            \vspace{0.02\linewidth}
            \scriptsize{\textbf{\textsf{Inconsistent Chroma}}}
        \end{center}
    \end{minipage}
    \hfill
    \begin{minipage}[b]{0.193\linewidth}
        \vspace{0.10\linewidth}
        \begin{center}
            %inconsistent hue
            \includegraphics[width=\textwidth]{\mm{39880}}\\
            \vspace{0.024cm}
            \includegraphics[width=\textwidth]{\mm{38588}}\\
            \vspace{0.024cm}
            \includegraphics[width=\textwidth]{\mm{39572}}\\
            \vspace{0.024cm}
            \includegraphics[width=\textwidth]{\mm{39824}}\\
            \vspace{0.024cm}
            \includegraphics[width=\textwidth]{\mm{40597}}\\
            \vspace{0.024cm}
            \includegraphics[width=\textwidth]{\mm{41006}}\\
            \vspace{0.024cm}
            \includegraphics[width=\textwidth]{\mm{40629}}\\
            \vspace{0.024cm}
            \includegraphics[width=\textwidth]{\mm{41943}}\\
            \vspace{0.02\linewidth}
            \scriptsize{\textbf{\textsf{Inconsistent Hue}}}
        \end{center}
    \end{minipage}
    \hfill
    \begin{minipage}[b]{0.193\linewidth}
        \vspace{0pt}
        \begin{center}
            % color polution (usually closeups)
            \includegraphics[width=\textwidth]{\mm{39090}}\\
            \vspace{0.014cm}
            \includegraphics[width=\textwidth]{\mm{40353}}\\
            \vspace{0.014cm}
            \includegraphics[width=\textwidth]{\mm{39332}}\\
            \vspace{0.014cm}
            \includegraphics[width=\textwidth]{\mm{37403}}\\
            \vspace{0.014cm}
            \includegraphics[width=\textwidth]{\mm{40124}}\\
            \vspace{0.014cm}
            \includegraphics[width=\textwidth]{\mm{38790}}\\
            \vspace{0.014cm}
            \includegraphics[width=\textwidth]{\mm{43015}}\\
            \vspace{0.014cm}
            \includegraphics[width=\textwidth]{\mm{43407}}\\
            \vspace{0.02\linewidth}
            \scriptsize{\textbf{\textsf{Edge Pollution}}}
        \end{center}
    \end{minipage}
    \hfill
    \begin{minipage}[b]{0.193\linewidth}
        \vspace{0pt}
        \begin{center}
            % color bleeding
            \includegraphics[width=\textwidth]{\mm{38463}}\\
            \vspace{0.055cm}
            \includegraphics[width=\textwidth]{\mm{44141}}\\
            \vspace{0.055cm}
            \includegraphics[width=\textwidth]{\mm{41892}}\\
            \vspace{0.055cm}
            \includegraphics[width=\textwidth]{\mm{43889}}\\
            \vspace{0.055cm}
            \includegraphics[width=\textwidth]{\mm{41901}}\\
            \vspace{0.055cm}
            \includegraphics[width=\textwidth]{\mm{37605}}\\
            \vspace{0.055cm}
            \includegraphics[width=\textwidth]{\mm{42051}}\\
            \vspace{0.02\linewidth}
            \scriptsize{\textbf{\textsf{Color Bleeding}}}
        \end{center}
    \end{minipage}
    \end{center}
    \caption{\small
        \textbf{Failure cases.}
        Examples of the five most common failure cases: the whole image lacks saturation ({\em Too Desaturated}); inconsistent chroma in objects or regions, causing parts to be gray ({\em Inconsistent Chroma}); inconsistent hue, causing unnatural color shifts that are particularly typical between red and blue ({\em Inconsistent Hue}); inconsistent hue and chroma around the edge, commonly occurring for closeups where background context is unclear ({\em Edge Pollution}); color boundary is not clearly separated, causing color bleeding ({\em Color Bleeding}).
    }
    \label{fig:ctest10k-supp-failures1}
\end{figure}

\section{Document changelog}
\label{sec:changelog}
Overview of document revisions:

\begin{itemize}
    \item[\textbf{v1}] Initial release. \\
    \item[\textbf{v2}] ECCV 2016 camera-ready version. Includes discussion about concurrent work and new experiments using colorization to learn visual representations (Section~\ref{sec:representation-learning}). \\
    \item[\textbf{v3}] Added overlooked reference.
\end{itemize}

%\end{appendices}

\end{document}